\documentclass{article} % For LaTeX2e
\usepackage{iclr2026_conference,times}

\usepackage{amsmath,amsfonts,bm}

\def\eqref#1{equation~\ref{#1}}
\def\1{\bm{1}}

\DeclareMathAlphabet{\mathsfit}{\encodingdefault}{\sfdefault}{m}{sl}
\SetMathAlphabet{\mathsfit}{bold}{\encodingdefault}{\sfdefault}{bx}{n}

\usepackage{hyperref}
\usepackage{url}
\usepackage{graphicx}
\usepackage{sidecap}
\usepackage{fontawesome}

\title{From Memories to Maps: Mechanisms of In-Context Reinforcement Learning in Transformers}

\author{Ching Fang\\
Department of Neurobiology,\\
Harvard Medical School \&\\
Kempner Institute, Harvard University,\\
Boston, MA, USA. \\
\texttt{chingfang17@gmail.com}
\And
Kanaka Rajan\footnote[1]{}\\
Department of Neurobiology,\\
Harvard Medical School \&\\
Kempner Institute, Harvard University,\\
Boston, MA, USA. \\
\texttt{kanaka\_rajan@hms.harvard.edu}\thanks{corresponding author: Kanaka Rajan}
}

\iclrfinalcopy % Uncomment for camera-ready version, but NOT for submission.
\begin{document}

\maketitle

\begin{abstract}
Humans and animals show remarkable learning efficiency, adapting to new environments with minimal experience. 
This capability is not well captured by standard reinforcement learning algorithms that rely on incremental value updates.
Rapid adaptation likely depends on episodic memory---the ability to retrieve specific past experiences to guide decisions in novel contexts.
Transformers provide a useful setting for studying these questions because of their ability to learn rapidly in-context and because their key-value architecture resembles episodic memory systems in the brain.
We train a transformer to in-context reinforcement learn in a distribution of planning tasks inspired by rodent behavior.
We then characterize the learning algorithms that emerge in the model.
We first find that representation learning is supported by in-context structure learning and cross-context alignment, where representations are aligned across environments with different sensory stimuli.
We next demonstrate that the reinforcement learning strategies developed by the model are not interpretable as standard model-free or model-based planning.
Instead, we show that in-context reinforcement learning is supported by caching intermediate computations within the model's memory tokens, which are then accessed at decision time.
Overall, we find that memory may serve as a computational resource, storing both raw experience and cached computations to support flexible behavior.
Furthermore, the representations developed in the model resemble computations associated with the hippocampal-entorhinal system in the brain, suggesting that our findings may be relevant for natural cognition.
Taken together, our work offers a mechanistic hypothesis for the rapid adaptation that underlies in-context learning in artificial and natural settings.
\end{abstract}

\section{Introduction}

Animals can learn efficiently and rapidly adapt to new environments with minimal experience. 
For example, humans can infer underlying structure or learn new concepts from just a handful of examples and mice in maze tasks can identify optimal paths after only a few successful trials \citep{meister2022learning}. 
Standard reinforcement learning (RL) algorithms, which typically rely on incremental value updates to shape decision-making, does not capture this rapid learning behavior well \citep{eckstein2024hybrid}.
One explanation is that biological agents possess useful priors shaped by evolution and experience, allowing them to generalize quickly in naturalistic settings.
They also rely on episodic memory, the ability to recall specific past experiences to guide decisions in novel situations.
%Understanding how episodic memory contributes to rapid learning remains an open question in both neuroscience and machine learning.

Here, we ask how episodic memory operates not just as storage, but as a computational substrate for rapid learning and decision-making. We train a transformer model to perform in-context reinforcement learning \citep{lee2023supervised} on navigation tasks inspired by rodent behavior.
In each new environment, the model receives exploratory trajectories as context and infers a goal-directed policy.
Transformers are especially relevant not only because of their established capabilities for rapid and flexible in-context learning \citep{dong2022survey, brown2020language, lampinen2024broader}, but also because their key–value memory architecture has been linked to models of episodic memory in the brain \citep{krotov2020large, tyulmankov2021biological, fang_2025}.
Understanding the reinforcement learning strategy that emerges in these models can provide new hypotheses for how memory-based computations might support flexible decision-making in new environments.

\begin{figure}[t!]
  \centering
  \includegraphics[width=\textwidth]{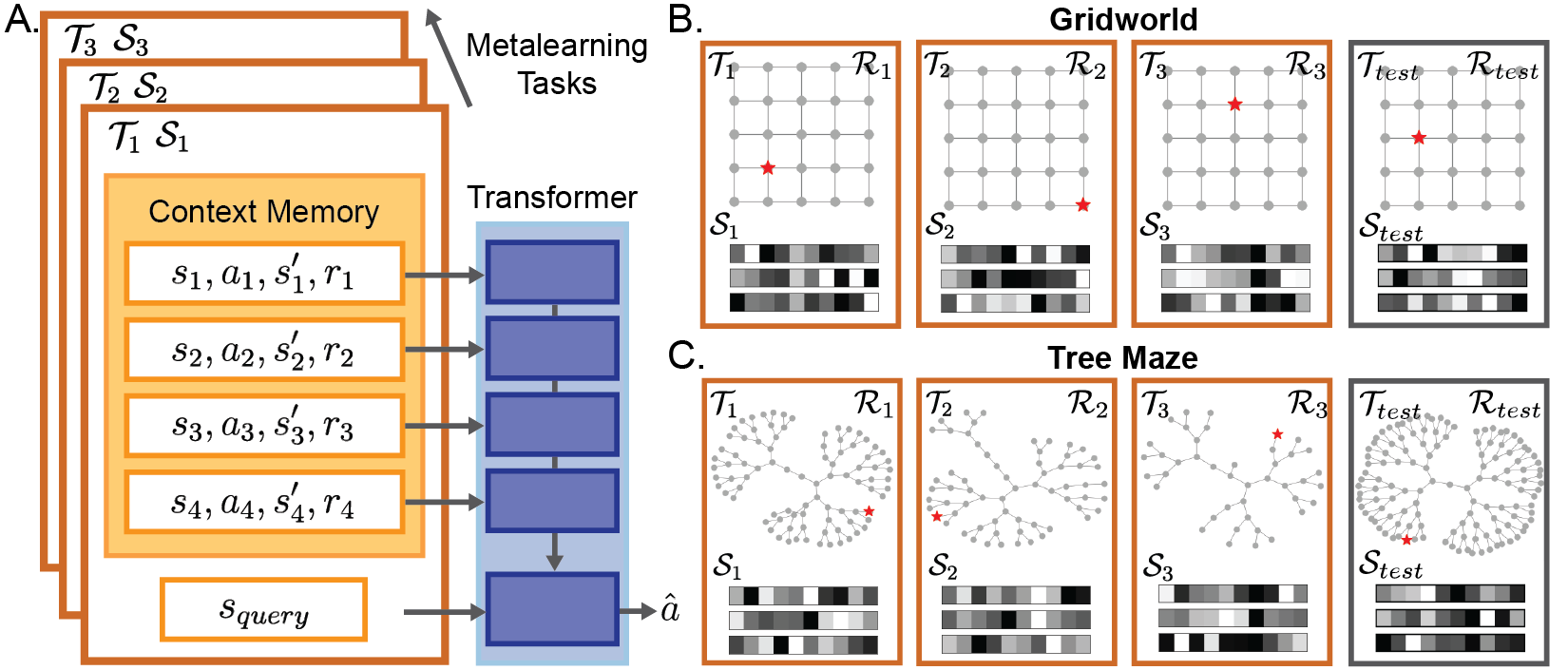}
\caption{\textbf{A transformer is trained to in-context reinforcement learn in diverse planning tasks.}
\textbf{A.} Diagram of meta-learning setup.
% Each task in the training distribution is defined by a state transition function $\mathcal{T}$, a state encoding function $\mathcal{S}$, and a reward function $\mathcal{R}$. 
For each task, the model is trained via supervision to predict the optimal action from a query state $s_{\text{query}}$, given memories of RL transition tuples sampled in-context. 
\textbf{B.} Illustration of three training tasks (orange) and one test task (gray) from the gridworld distribution. In each task, the underlying graph structure is fixed, but the reward location (red star) can vary. Each state is encoded as a random Gaussian vector (bottom). Importantly, test task state encodings are novel.
\textbf{C.} As in (B), but for the tree maze distribution. The training set graph structures are drawn from probabilistically branching trees, while the test set structure is a full binary tree.
}
  \label{fig:1}
\end{figure}

We focus on two task suites: spatially regular gridworlds and hierarchically structured tree mazes (Fig \ref{fig:1}BC). 
While both require memory to support goal-directed behavior, they differ sharply in geometry: gridworlds are Euclidean and spatially continuous, whereas tree mazes are non-Euclidean and branch-structured.
This contrast allows us to evaluate how learned in-context strategies generalize across structural regimes known to challenge standard sequence models-- for instance, language models are known to struggle with symbolic reasoning and hierarchical generalization in tree-structured domains \citep{bogin2022probing, ruiz2021graph, keysers2020measuring}.

In this paper, we make the following contributions:
\begin{itemize}
    \item We show that transformers trained to in-context reinforcement learn develop consistent representation learning strategies: structure learning within contexts, and alignment of representations across contexts with shared regularities.
    \item We demonstrate that the model learns computations found in natural cognition. Representation learning strategies are consistent with suggested roles for hippocampus and entorhinal cortex, and memory recall patterns at decision time are consistent with hippocampal replay.
    \item We give descriptions for the in-context RL algorithms that emerge. We show with mechanistic analysis that the model does not use standard model-free or model-based RL methods. Instead, strategies tend to rely on intermediate computations cached in memory tokens, demonstrating how episodic memory can be used as a computational workspace.
\end{itemize}

%Overall, we find that meta-learning produces agents capable of rapid adaptation using memory-based strategies. Notably, the model’s internal representations resemble those observed in neural recordings, despite its abstract architecture \citep{stachenfeld2017hippocampus, pfeiffer2013hippocampal, chettih2024barcoding}. These findings suggest that explaining rapid learning may require expanding our theoretical toolkit beyond standard RL paradigms. We propose that episodic memory should be understood not only as storage, but as a substrate for computation.

% Diagram of the meta-learning setup.

%\section{Related Works}
\paragraph{Related works}  In meta-RL settings, the outer learning loop shapes the weights of the network to learn an  algorithm that can be deployed in-context. The in-context learning within each task occurs via activation dynamics-- through memory and internal state-- rather than parameter updates \citep{beck2023survey, sandbrink2024modelling, lampinen2024broader}. 
Early examples of this approach include RL$^2$, which meta-trains a recurrent neural network using RL in the outer loop, such that an in-context RL strategy emerges in the inner loop \citep{wang2016learning, duan2016rl}.
Subsequent work extended this approach to include explicit episodic memory mechanisms, combining RNNs with key–value memory architectures \citep{ritter2018been, ritter2020rapid, team2023human}. 
More recently, \citet{lee2023supervised} proposed decision-pretrained transformers (DPTs), which use supervised training in the outer loop to induce in-context RL behavior in the inner loop.
We adopt DPTs in this work both for practical reasons--their scalability and ease of training--and for scientific ones: the transformer architecture allows us to probe how memory-based computation supports in-context learning. This latter motivation is inspired by recent findings suggest that key–value architectures, like the transformer, offer a useful computational analogy for episodic memory \citep{krotov2020large, ramsauer2020hopfield, tyulmankov2021biological, whittington2021relating, fang_2025, chandra2025episodic}. See App. \ref{sec:appendix-related-work} for a more thorough discussion of related work.

\section{Experimental Methods}
\textbf{Meta-learning procedure}\quad
We adopt the decision-pretraining framework from \citet{lee2023supervised} (Fig.~\ref{fig:1}A), training models via supervision to learn optimal policies from in-context experience. Each task is a Markov Decision Process defined by state encoding function $\mathcal{S}$, action space $\mathcal{A}$, transition function $\mathcal{T}$, reward function $\mathcal{R}$. For each task, the model receives an in- context dataset $\mathcal{D}$ of RL transition tuples $(s, a, s', r)$ gathered from an exploratory policy, plus a query state $s_{\text{query}} \in \mathcal{S}$. The model is meta-trained to predict the action from an oracle policy. At test time, the model generalizes to held-out tasks with novel sensory observations using only in-context information, demonstrating in-context reinforcement learning. We focus on offline settings where $\mathcal{D}$ comes from random exploration, and more details can be found in App. \ref{sec:appendix-task-construction}.

\textbf{Structure of task suites}\quad
Our first task distribution is a $5 \times 5$ gridworld in which the reward location is fixed but hidden from the agent. 
This setting is loosely inspired by the Morris water maze, a behavioral task used to study how animals use memory to navigate unknown environments \citep{vorhees2006morris}.
Across tasks, $\mathcal{T}$ is fixed, while $\mathcal{S}$ and $\mathcal{R}$ vary (Fig.~\ref{fig:1}B).
At test time, the model is deployed in a gridworld with novel sensory observations and a new reward location. Our second task distribution consists of tree-structured mazes, which introduce hierarchical state transitions and sparse rewards. These are settings where rodents have been shown to display rapid learning \citep{rosenberg2021mice}.
The meta-training set consists of binary trees generated with some branching probability so that $\mathcal{T}$ varies across tasks (Fig.~\ref{fig:1}C).
% Invalid transitions (e.g., from the root to a parent) result in the agent remaining in place.
$\mathcal{S}$ and $\mathcal{R}$ also vary across tasks.
The action space consists of four options: stay, move to the parent node, or move to either child node.
At test time, the model is evaluated on a full 7-layer binary tree, consistent with \citet{rosenberg2021mice}, again using novel state encodings not seen during training.

In both tasks, states are represented by 10-dimensional random vectors.
%, with spatial correlations introduced across states.
% The in-context dataset $\mathcal{D}$ is generated using a random walk policy with additional heuristics.
Full task details are in App. \ref{sec:appendix-task-construction}.
Together, these two tasks allow us to analyze model behavior in spatially regular environments and branching tree structures (which may be a relevant analogy for language generation tasks).

\textbf{Model architecture and selection}\quad
Our base architecture is a causal, GPT2-style transformer with 3 layers and 512-dimensional embeddings.
We provide the context memory $\mathcal{D}$ before the query token $s_{query}$.
This ordering supports an interpretation in which previous experiences are stored as cached key–value memories that are retrieved at query, or decision-making, time.
% A useful analogy is the autoregressive interpretation of linear attention transformers \citep{katharopoulos2020transformers}.
% This contrasts with \citet{lee2023supervised}, where the query appears first.
% To efficiently train the model with this token ordering, we implement custom attention masks during training (Appendix~\ref{sec:appendix-model-training}).
% We meta-train five models from different random seeds and select the best-performing one for analysis.
Additional details on architecture and training are in App.~\ref{sec:appendix-model-training}. We also test alternative modeling choices in App.~\ref{sec:appendix-results-sensitivity}.

\begin{figure}[t!]
  \centering
  \includegraphics[width=\textwidth]{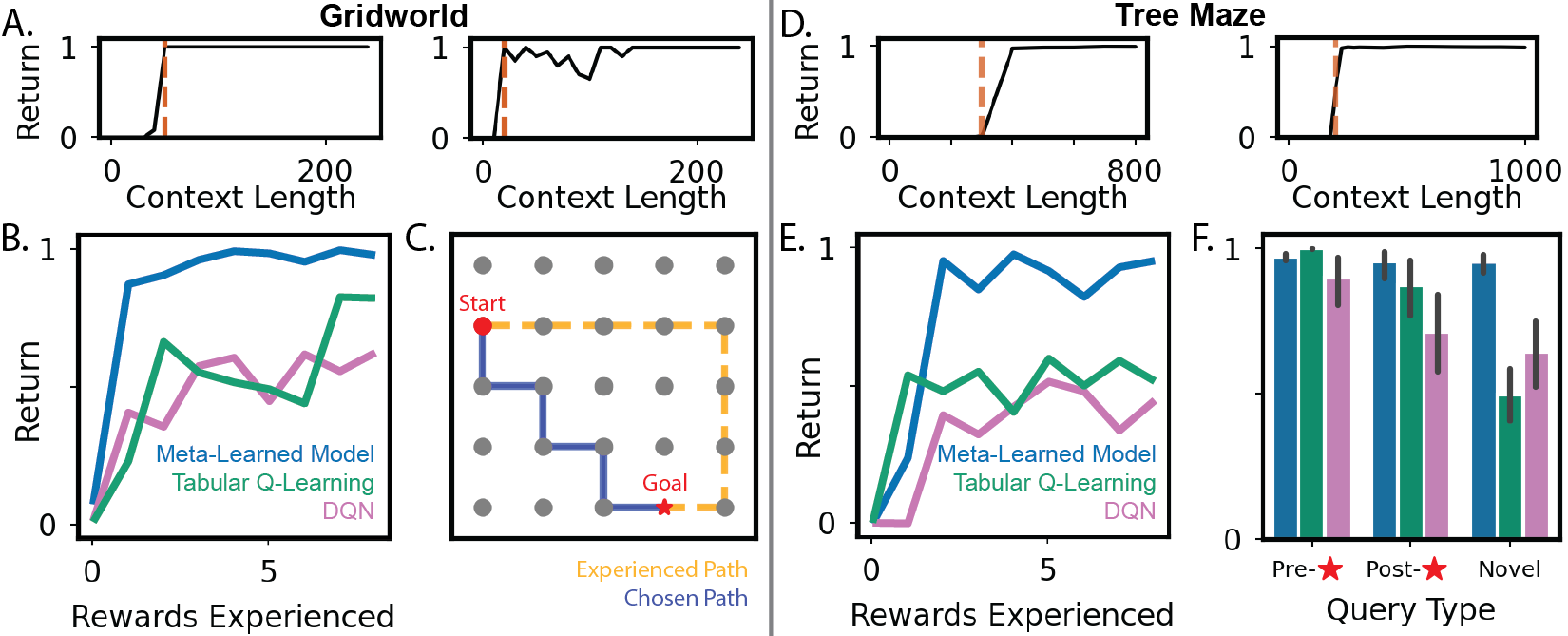}
\caption{\textbf{Transformers can rapidly learn and plan in new tasks.}
\textbf{A.} Average max-normalized return in two held-out gridworld environments as a function of context length. For each context length, $20$ query states are sampled with test horizon $15$.
\textbf{B.} As in (A), but return is plotted against the number of rewards experienced in-context and averaged over $50$ held-out environments. Blue: meta-learned transformer; Green: tabular Q-learning; Pink: DQN.
\textbf{C.} Example of shortcut behavior in a held-out gridworld. 
The model experiences a circuitous trajectory (orange), but can infer a more efficient path (blue).
\textbf{D., E.} As in (A, B), but for tree mazes and test horizon $100$.
\textbf{F.} As in (D,E), but shown only for context length $800$ and subdivided by query type: states seen before reward (Pre-$\star$), after reward (Post-$\star$), or never seen in context (Novel; not used in E).
Error bars show 95\% C.I.
}
  \label{fig:2}
\end{figure}

\section{Results}
\subsection{Transformers learn a RL strategy to rapidly solve planning tasks.}
We first evaluate the agent's performance in new environments with novel sensory observations.

\textbf{Gridworld} \quad
In held-out gridworld environments, we test the agent with query states that were observed in-context and located at least $6$ steps from the goal.
We plot return as a function of context length (Fig.~\ref{fig:2}A; App.~\ref{sec:appendix-fig_2}). 
% If no eligible query states meet the selection criteria, return is recorded as zero.
The meta-learned agent often navigates directly to the reward after a single exposure, mirroring one-shot learning reported in rodents navigating water mazes \citep{steele1999delay}.
To summarize test environment performance, we plot return as a function of the number of rewards experienced in-context (Fig.~\ref{fig:2}B, blue).
As expected from Fig.~\ref{fig:2}A, the agent achieves near-maximal performance after just one exposure, with only minor improvements thereafter.

We next compare the meta-learned model to standard reinforcement learning methods.
Specifically, we train a tabular Q-learning agent and a deep Q-network (DQN) on each test environment.
% The tabular agent is a useful benchmark for performance when the challenge of learning representations is removed.
Each Q-learning agent is trained using a replay buffer containing the same in-context dataset $\mathcal{D}$ provided to the meta-learned model (App.~\ref{sec:appendix-q-learning}).
We again summarize performance  in Fig.~\ref{fig:2}B.
The performance gap between both Q-learning agents and the meta-learned model is substantial but expected, reflecting the utility of meta-learned priors.
The advantage of the tabular agent over DQN demonstrates that representation learning adds additional difficulty in novel environments.
Overall, describing the learning efficiency seen in animals may require moving beyond single-task RL frameworks.

Finally, we observe that the meta-learned agent discovers shortcut paths to reward.
Even when the agent only observes a circuitous path in-context, it infers a policy that selects the shortest route to reward—often through previously unseen states (Fig.~\ref{fig:2}C).
Quantitatively, the model selects shortcut paths in over 60\% of test simulations (App.~\ref{sec:appendix-fig_2}).
This suggests that the agent has internalized the Euclidean geometry of the environment, a feature that we analyze more deeply in later sections.

\textbf{Tree Mazes}\quad We next evaluate the meta-learned agent in test tree mazes, where the agent rapidly learns the task after only a few reward exposures (Fig.~\ref{fig:2}DE; App.~\ref{sec:appendix-fig_2}).
As before, the meta-learned agent captures rapid learning more effectively than Q-learning baselines (Fig.~\ref{fig:2}E).
The tabular agent again outperforms the DQN, confirming that representation learning remains a core challenge.

To gain insight into the priors acquired through meta-learning, we evaluate all models at a long context length ($800$ timesteps).
We stratify performance by the type of query state (Fig.~\ref{fig:2}F).
When the query state had already been seen prior to any reward, the tabular agent performed comparably to the meta-learned model.
However, when the query state was seen only after the final reward, the tabular agent underperforms. In standard Q-learning, states encountered only along paths away from reward do not receive value propagation.
This suggests that the meta-learned model acquires a useful prior: the ability to infer inverse actions.
Finally, we evaluate performance when the query state was never encountered during context.
In this setting, the meta-learned agent performs better than both Q-learning baselines, likely due to a learned prior over action selection.

\begin{figure}[t!]
  \centering
  \includegraphics[width=\textwidth]{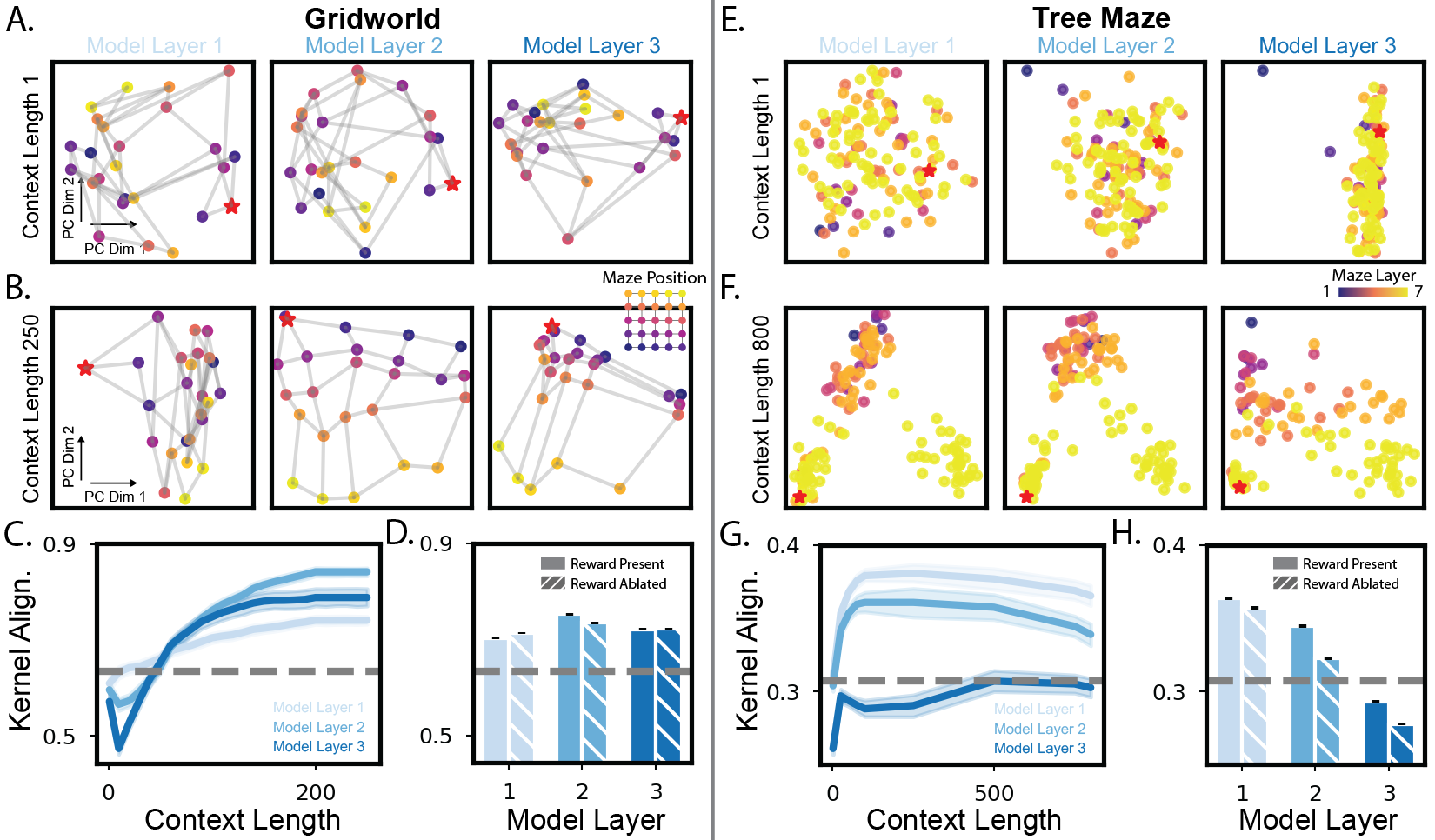}
  \caption{\textbf{Model representations are shaped by in-context structure learning.} 
\textbf{A.} An example test gridworld. Query token representations are visualized for each state after projection onto the first two principal components, for layer 1 (left), layer 2 (middle), and layer 3 (right). Context length is 1. Points are colored by graph location; gray lines indicate true connectivity. Reward is marked with a red star. 
\textbf{B.} As in (A), but with context length 250.
\textbf{C.} Kernel alignment between model representations and latent graph structure as a function of context length, across $100$ environments. Shading shows 95\% C.I.; colors denote model layer. Dashed line shows baseline from raw inputs.
\textbf{D.} As in (C), but for context length 250, with reward ablation (shaded bars).
\textbf{E, F.} As in (A, B), but for test tree mazes. Points are colored by maze depth.
\textbf{G, H.} As in (C, D), but for test tree mazes.}
  \label{fig:3}
\end{figure}

%\subsection{The RL strategy learned by the model relies on in-context structure learning to shape representations.}
\subsection{RL strategy shapes representations via in-context structure learning.}
\label{sec:in-context}
The behavioral results in Fig.~\ref{fig:2}B,E highlight that representation learning poses a key computational challenge in these tasks.
How does the model organize state representations—and does a structured representation learning strategy emerge during in-context processing?

\textbf{Gridworld}\quad
We begin by visualizing low-dimensional embeddings of model activity across gridworld states.
Because the query token represents the agent’s current state, we extract its activity as the basis for representation analysis.
We project the $512$-dimensional query representations into a 2D PCA space at each model layer, for both short and long context lengths (Fig.~\ref{fig:3}A--B; App.~\ref{sec:appendix-in-context}).
% Each point is colored by the spatial location of the corresponding query state in the gridworld.
With limited context, the model’s representations are disorganized and show no spatial structure (Fig.~\ref{fig:3}A).
As in-context experience increases, the representations become more structured and reflect the latent geometry of the gridworld (Fig.~\ref{fig:3}B).
This structure resembles predictive representation learning, but crucially, no such objective was imposed during training.

To quantify this, we compute the kernel alignment between model representations and the latent environment structure across held-out tasks (Fig.~\ref{fig:3}C; App.~\ref{sec:appendix-kernel-alignment}).
Kernel alignment increases with context length, with layer 2 consistently exhibiting the strongest correspondence to latent structure.
Surprisingly, representation structure is largely unaffected by the presence of reward (Fig.~\ref{fig:3}D; App.~\ref{sec:appendix-in-context},~\ref{sec:appendix-kernel-alignment}). Overall, we find that in-context experience induces geometry-aligned state representations.

% Overall, we find that in-context experience in gridworld induces geometry-aligned state representations.
% the presence or absence of reward

% We next ask whether in-context representation learning also emerges in agents trained on the tree maze task.

\textbf{Tree Mazes}\quad
Does in-context representation learning also emerge in agents trained on the tree maze task? We repeat the PCA projection analysis on query token representations in tree mazes (Fig.~\ref{fig:3}E--F; App.~\ref{sec:appendix-in-context}).
As context increases, representations organize into a bifurcating structure that roughly mirrors the maze’s hierarchical layout (Fig.~\ref{fig:3}F), which suggests that the model learns coarse, high-level structure rather than fine-grained spatial layout.
Consistent with this intuition, kernel alignment also increases with context length in tree mazes (Fig.~\ref{fig:3}G), but remains lower than in the gridworld task.
Representations in the tree maze are more strongly modulated by reward (Fig.~\ref{fig:3}H; App.~\ref{sec:appendix-in-context},~\ref{sec:appendix-kernel-alignment}).

% This aligns with the visualizations in Fig.~\ref{fig:3}F, which suggest that the model learns coarse, high-level structure rather than fine-grained spatial layout.

Taken together, these results indicate that the model meta-learns in-context representation strategies that vary in granularity across task domains.
This provides \textbf{normative support for the hypothesis that structure learning facilitates efficient decision-making}, and in fact in-context structure learning has also been found in the representations of large language models \citep{park2024iclr}.
This observation parallels predictive map formation in the hippocampus \citep{stachenfeld2017hippocampus}—long hypothesized as a computational scaffold for memory—and suggests that similar principles can emerge in artificial agents through meta-learning.

\begin{figure}[t!]
  \centering
  \includegraphics[width=\textwidth]{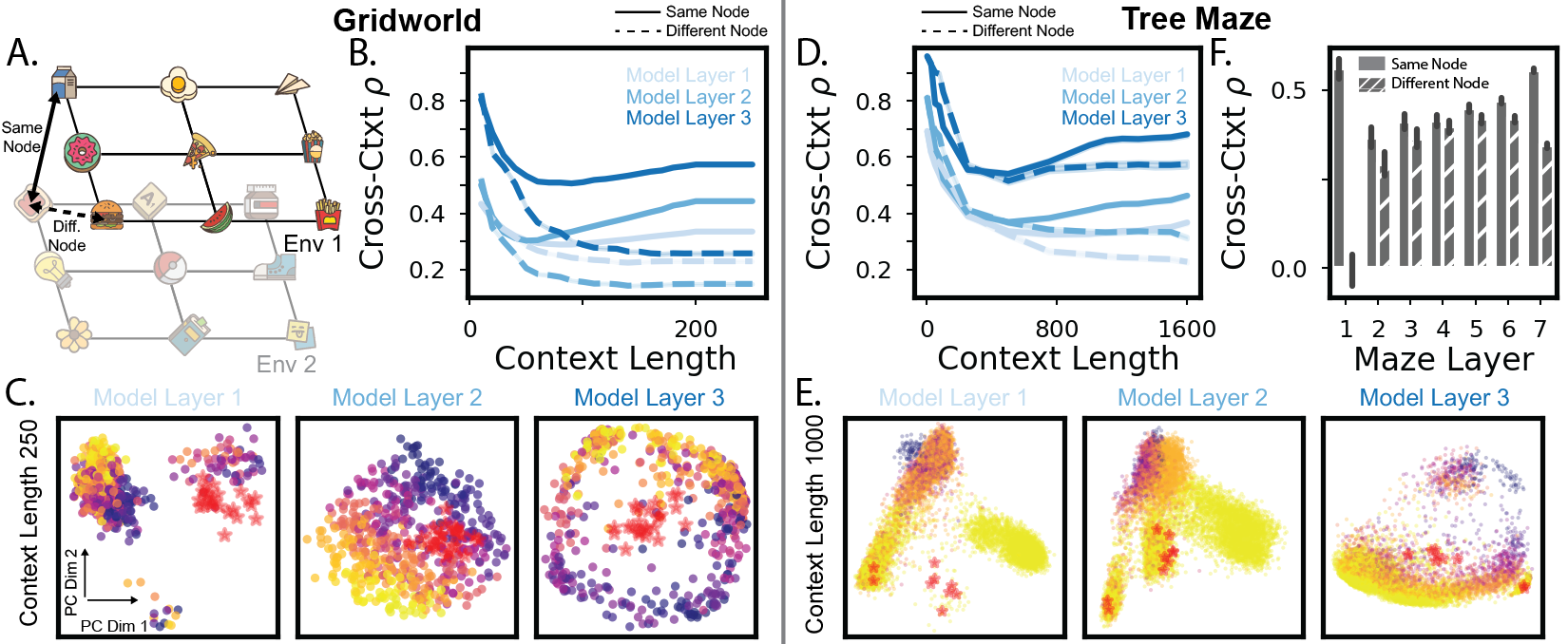}
\caption{\textbf{As context grows, representations across environments with similar structure are aligned.}
\textbf{A.} Diagram of cross-environment alignment \citep{whittington2020tolman}. Although sensory inputs differ, environments share latent structure, and representations of matching latent states should be similar.
\textbf{B.} Average pairwise Pearson correlation coefficient of node representations across $100$ gridworld environments, as a function of context length. Solid lines: same-node comparisons. Dashed lines: different-node comparisons. Shading shows 95\% C.I.. Line color denotes model layer.
\textbf{C.} PCA visualization of representations pooled from 15 randomly selected visualizations.
\textbf{D.} As in (B), but for $50$ tree mazes.
\textbf{E.} As in (C), but for tree mazes.
\textbf{F.} Summary of (D) at context length 1600, averaged across layers. X-axis denotes the maze-layer of the comparison node.
}
  \label{fig:4}
\end{figure}

%\subsection{Representation learning also aligns representations across environments with shared structure.}
\subsection{Representations are reused across environments with shared structure.}
\label{sec:cross-context}
In neuroscience, the hippocampal--entorhinal circuit is thought to support structure learning across contexts \citep{buckmaster2004entorhinal, kumaran2009tracking, whittington2020tolman}.
A leading hypothesis holds that the hippocampus encodes context-specific experiences, while the entorhinal cortex abstracts shared structure across environments \citep{whittington2020tolman}.
Do similar cross-context alignment strategies emerge in meta-learned agents?

% A key signature of cross-context structure learning is the alignment of internal representations across environments with shared topology.
% Specifically, even when sensory observations differ across environments, states occupying the same grid location should be encoded more similarly than states from different locations (Fig.~\ref{fig:4}A).

\textbf{Gridworld}\quad
A key signature of cross-context structure learning is the alignment of internal representations across environments with shared topology: even when sensory observations differ across environments, states occupying the same grid location should be encoded more similarly than states from different locations (Fig.~\ref{fig:4}A).
To test this, we compute pairwise correlations between model representations of the same graph node across different test environments (Fig.~\ref{fig:4}B).
We separate correlation scores by whether the compared states occupy the same or different graph nodes.
With limited context, representations appear collapsed—showing high correlation across all states regardless of node identity.
As context length increases, this gap widens: states from the same node become more aligned than those from different nodes (Fig.~\ref{fig:4}B; solid vs.\ dashed).

This alignment is also visually apparent when directly inspecting the model’s internal representations.
We aggregate representations from 50 test environments and project them into a shared 2D PCA space (Fig.~\ref{fig:4}C).
In layers 2 and 3, representations from corresponding graph nodes cluster across environments, reflecting shared latent structure.
Notably, representations of goal states also align across environments, despite the reward location being randomized.

\textbf{Tree Mazes}\quad
We repeat the same analyses in tree mazes and find similar alignment strategies emerge  (Fig.~\ref{fig:4}DE).
We further analyze the correlation scores by the node position within the tree (Fig.~\ref{fig:4}F).
Cross-context alignment is strongest for states near the root or leaves of the tree.
This is consistent with \S \ref{sec:in-context}, where the representations capture coarse structure over precise positional detail in tree mazes.
Importantly, neither in-context nor cross-context representation alignment was explicitly trained—these strategies emerge opportunistically as a byproduct of meta-learning.

\begin{figure}[t!]
  \centering
  \begin{minipage}[c]{0.427\textwidth}
    \includegraphics[width=\textwidth]{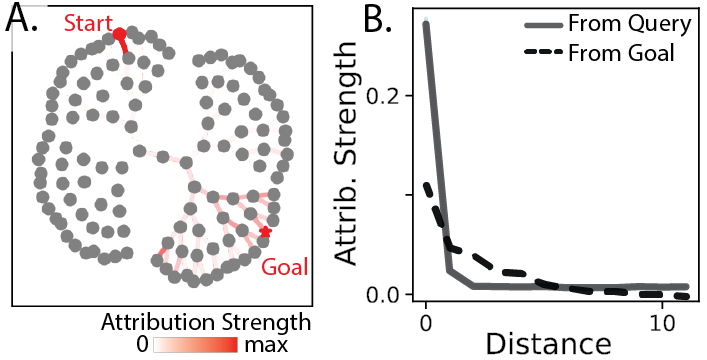}
  \end{minipage}\hfill
  \begin{minipage}[c]{0.55\textwidth}
\caption{\textbf{Memory retrieval at decision time shows limited expansion from the query state and the goal state.}
  \textbf{A.} Example tree maze environment, context length 800. Edge color indicates gradient attribution strength for each transition.
  \textbf{B.} Average attribution strength of each context memory vs distance from query state and goal state, across 50 environments.
  }
  \label{fig:5}
  \end{minipage}
\end{figure}

%\subsection{The RL strategy that emerges from meta-learning does not show signatures of value-based learning or model-based planning.}

\subsection{RL strategy is neither value-based learning nor model-based planning.}
\label{sec:not-mf-mb}
Thus far, we have shown that meta-learned agents replicate the rapid learning dynamics observed in animal behavior.
A key component of this success is the emergence of structured representations from contextual input.
We now turn to characterizing the mechanisms of the underlying RL strategy.
We begin by testing whether the model exhibits hallmarks of standard model-free reinforcement learning.
We test whether value information can be linearly decoded from model representations, but overall did not find evidence for this (Apps. \ref{sec:appendix-linear-decoder} and \ref{sec:appendix-model-free}).

Next, we test whether the model exhibits hallmarks of standard model-based RL, which typically requires path planning from query to goal.
Planning need not follow a strictly forward rollout, and transformers in particular can implement diverse state-tracking strategies \citep{li2025language}.
Critically, however, all such strategies depend on retrieving intermediate states along the path from query to goal during decision time.
To evaluate this, we assess which context-memory tokens influence the model’s decision at a given query state.
Using integrated gradients, we measure attribution strength for memory tokens along the query$\rightarrow$goal path (Fig.~\ref{fig:5}; App.~\ref{sec:appendix-model-based}).
In both tasks, only tokens near the query and goal states show high attribution. As a further test, we also conduct attention ablations and find similar results (App.~\ref{sec:appendix-model-based}). 
Both results are inconsistent with path planning, which requires attending to transitions along the full route at decision time.

In summary, we find that the agent does not use value gradients or path planning to make decisions. We suggest that the learned in-context strategy lies outside the standard taxonomy of model-free and model-based reinforcement learning. Our analyses also reveal an additional neural prediction: memory retrieval at decision time should be biased toward experiences near the agent’s current location and its goal.
Such replay patterns have been observed in the hippocampus during spatial decision-making tasks \citep{jackson2006hippocampal, pfeiffer2013hippocampal, mattar2018prioritized}.

\begin{figure}[t!]
  \centering
  \includegraphics[width=\textwidth]{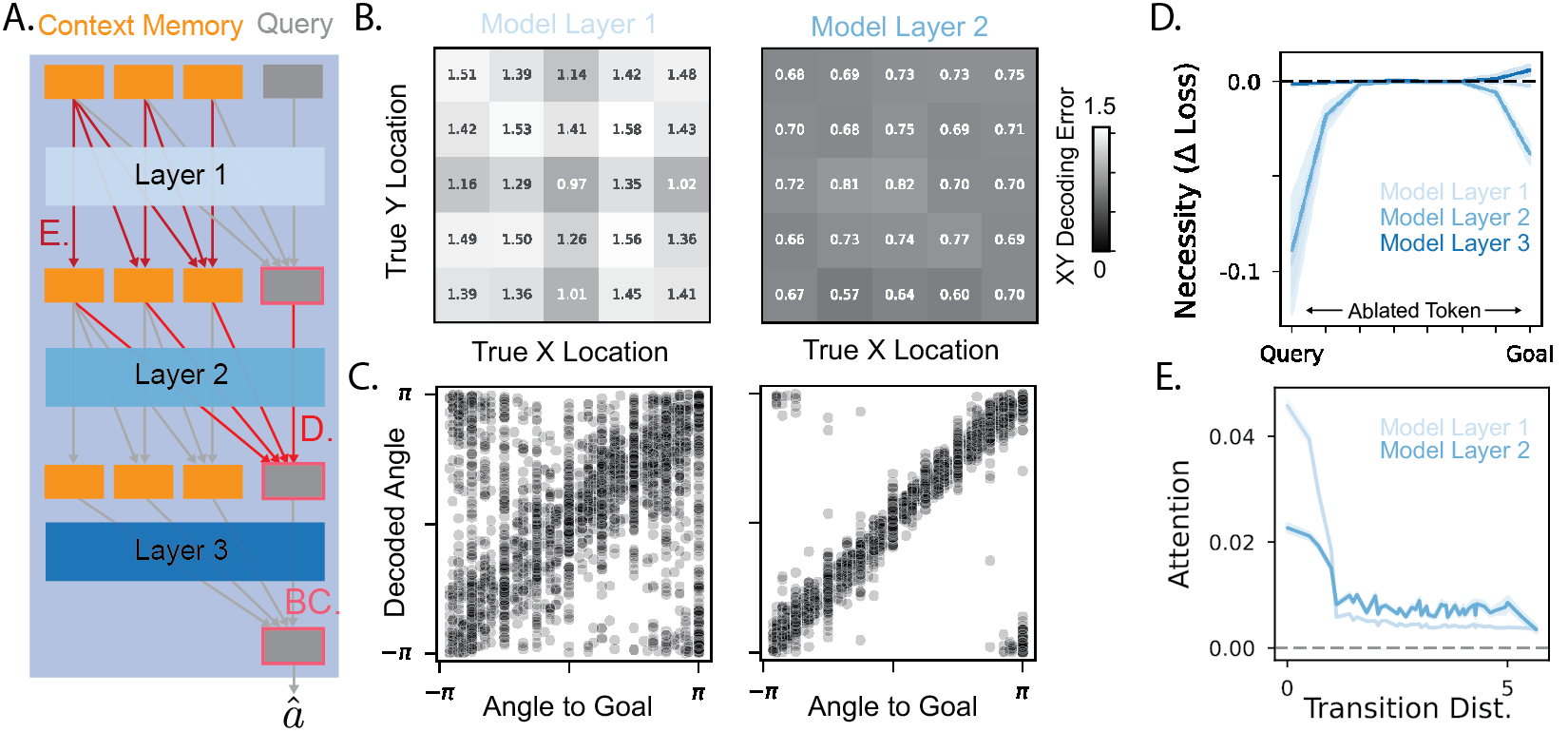}
\caption{\textbf{Gridworld tasks are solved by aligning internal representations to Euclidean space.}  
\textbf{A.} Overview of analysis steps.
\textbf{B.} Total XY decoding error across $60$ test environments, plotted by true XY position of the query state, given query token embeddings from layer 1 (left) and layer (right). Context length = 250. 
\textbf{C.} As in (B), but for decoded vs. true angle from query state to the goal.  
\textbf{D.} Change in cross-entropy loss after ablating context tokens along the query–goal path, plotted by ablated token position. 
Line color indicates layer of intervention. 
\textbf{E.} Average attention score between context-memory tokens as a function of spatial distance between tokens, for layers 1 and 2. (D,E) show mean over 50 environments, with shading for 95\% C.I.}
  \label{fig:6}
\end{figure}

%\subsection{Transformers learn diverse in-context RL strategies where intermediate computations are stored in context-memory tokens.}

\subsection{Models learn strategy where intermediate computations are stored in context-memory tokens.}
% \subsection{Intermediate computations are stored in context-memory tokens.}
\label{sec:mechanisms}
We now aim to describe the algorithms used by the model to plan in each task. To do so, we first review the roles of query and memory tokens.
The query token encodes the agent’s current state, while memory tokens represent previously observed transitions.
During inference, the query token attends to memory tokens to integrate past experience into its policy computation.
Across layers, both query and memory tokens are updated with newly computed features, allowing memory to serve as an active computational substrate.
To reveal how computation unfolds, we will focus on understanding which tokens are critical in each layer and what information is contained in tokens.

\textbf{Gridworld}\quad
In gridworld tasks, we suggest that the following strategy is used by the model:\vspace{-0.1in}
\begin{enumerate}
 \itemsep-0.05em 
    \item Use in-context experience to align representations to Euclidean space.
    \item Given a query state, calculate the angle in Euclidean space between query and goal state.
    \item Use the calculated angle to select an action in that direction.
\end{enumerate}\vspace{-0.1in}
We arrived at this hypothesis by first identifying the task-relevant variables that can be linearly decoded from the query token at each layer (Fig.~\ref{fig:6}A).
We train a linear decoder to predict the underlying XY position of the query state from its embedding (App.~\ref{sec:appendix-linear-decoder}) and evaluate accuracy on held-out environments with novel sensory observations.
Decoding accuracy improves across layers, with spatial position becoming reliably recoverable by layer 2 (error $<1$; Fig.~\ref{fig:6}B).
Building on this spatial structure, we next decode the angle from the query state to the goal-- again, this information can be accurately decoded by layer 2 (Fig.~\ref{fig:6}C).
Interestingly, both XY position and angle-to-goal can be decoded from the embeddings of the context-memory transitions $(s, a, s', r)$ as well.

To localize where angle-to-goal information may be computed, we test which context-memory tokens are necessary for correct decisions. Using attention ablations, we show that model performance relies on attending to tokens near the query and goal states in layer 2 (Fig.~\ref{fig:6}D, App. \ref{sec:appendix-gridworld-mech}).
We propose that layer 2 extracts the internal XY coordinates from query and goal state tokens to compute the relative angle between them.
To understand how XY information arises, we examine how state representations evolve through context-to-context attention.

We show that, across layers, attention patterns between context memory tokens shift from localized to distributed (Fig.~\ref{fig:6}E, App. \ref{sec:appendix-gridworld-mech}), suggesting that the model first stitches transitions locally before constructing global structure. Overall, we suggest that the model organizes memory to reflect Euclidean structure and use that geometry to guide action selection.
This explains the model’s ability to take unseen shortcuts (Fig.~\ref{fig:2}C).

\begin{figure}[t!]
  \centering
  \includegraphics[width=\textwidth]{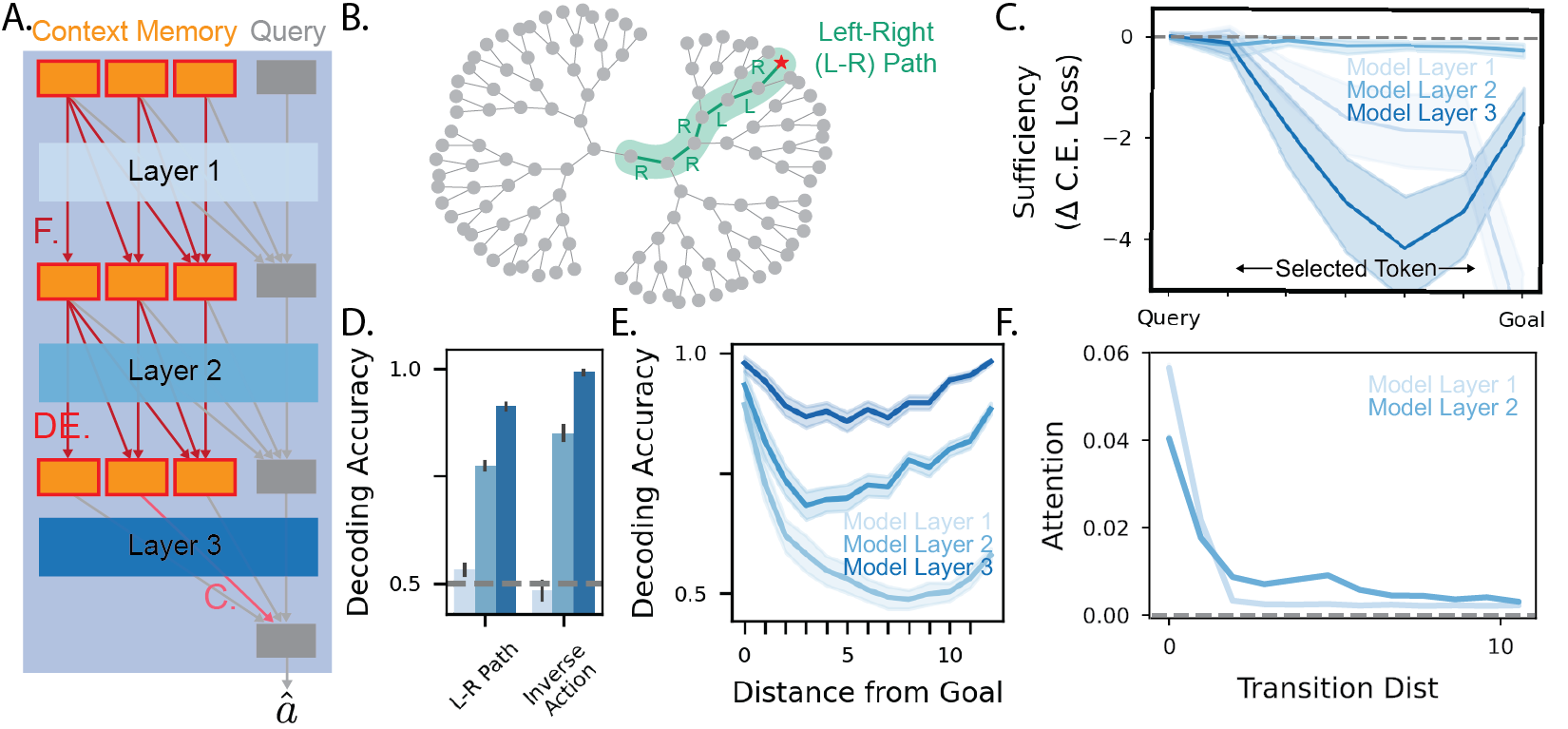}
  \caption{\textbf{Tree mazes are solved by tagging context-memory tokens on a critical path to reward.}  \textbf{A.} Overview of analysis steps. \textbf{B.} We focus on the path from root to goal (L-R path). \textbf{C.} Change in cross-entropy loss when attention is fixed to context-memory tokens at specific points on the L-R path. In all tests, the query state is the root. Line color indicates layer of intervention. Mean across $50$ environments, shading shows $95\%$ C.I. \textbf{D.} Given embeddings of context-memory transitions, the balanced accuracy of decoding their presence on the L-R path and the inverse action. Decoding the inverse action is only non-trivial for parent-node transitions so we test only on those. Embeddings are taken from the input tokens to each model layer (line color). Mean across $60$ environments, error bars show $95\%$ C.I. Dashed line is chance. \textbf{E.} L-R Path decoding accuracy from (D), but separated by how far each context-memory transition is from goal. \textbf{F.} As in Fig \ref{fig:6}E but for tree mazes.}
  \label{fig:7}
\end{figure}

\textbf{Tree Mazes}\quad
In tree mazes, a useful strategy can be to identify when the agent is on a critical path to reward and to default to the parent-node action otherwise.
 This is because there are only $6$ states in the maze (of $127$) where the optimal action is to transition to the left child or right child (Fig \ref{fig:7}B). These are the states on the path from root to reward, which we will call the left-right (L-R) path. Indeed, the model has a strong action bias to take parent-node transitions (App. \ref{sec:appendix-tree-maze-mech}). Overall, we find evidence that the model exploits this structure and learns the following strategy:\vspace{-0.1in}
 \begin{enumerate} 
 \itemsep-0.05em 
    \item Use in-context experience to stitch transitions backwards from the goal to root, and tag context-memory tokens that are on the L-R path.
    \item Given a query state, check if there are context-memory tokens that contain the query state and are on the L-R path. If not, default to taking the parent-node transition.
    \item Otherwise, extract the optimal action information from the tagged context-memory tokens. 
 \end{enumerate} \vspace{-0.1in}
We arrive at this hypothesis by asking how the model takes correct actions when it is on the L-R path. Again, we work backwards and analyze which context-memory tokens are sufficient to influence the output from the final model layer (Fig \ref{fig:7}C, App. \ref{sec:appendix-tree-maze-mech}). Surprisingly, we find that the model output is unaffected if the query token of the last layer attends only to context-memory tokens involving the query state. These tokens contain sufficient information for the model to make its decision.

With this in mind, our next question was to understand what information is contained in the context-memory tokens entering the last model layer. We repeat our linear decoding analyses on the context-memory tokens. Two variables are well-decoded. First, the inverse action for the transition represented in a context-memory token can be decoded with high accuracy (Fig \ref{fig:7}D). The other well-decoded variable is whether the context-memory token is a transition on the L-R path (Fig \ref{fig:7}D), regardless of direction (i.e., towards or away from goal). Possibly, at decision time the model tests if there are context-memory tokens that contain the query state and are tagged as being on the L-R path. If so, then the correct left/right action can be inferred from the same tagged tokens (in particular since inverse actions are also encoded). We find further evidence for this strategy by re-doing our sufficiency analysis from Fig \ref{fig:7}C with additional restrictions on the selected tokens (see App. \ref{sec:appendix-tree-maze-mech}). 

Finally, we ask how this information becomes present in the context-memory tokens. We plot the L-R path decoding accuracy from Fig \ref{fig:7}D by the distance from the context-memory token to goal (Fig \ref{fig:7}E). Across model layers, we see that the accuracy first improves for tokens closest to or farthest from the goal. Later, the accuracy improves for tokens at an intermediate distance from the goal, where more transition information must be integrated to know if the token is on the L-R path.
Furthermore, attention patterns between context memory tokens shift from localized to distributed across model layers (Fig \ref{fig:7}F).
Taken together, we suggest that path stitching occurs between context-memory tokens such that L-R path tokens are tagged expanding backwards from the goal.

\vspace{-1em}
\section{Conclusion}
% Write about limitations and future work.
% This metalearning framework gives a normative explanation for strategies that support rapid learning.
We have shown that rapid adaptation of agents in tasks relevant to natural cognition can be explained by RL strategies that lie outside traditional model-free or model-based frameworks. Despite this, our meta-learned model also displays phenomena expected from neural activity: learning of environment structure, alignment of representations across environments, and biased memory recall patterns at decision-time. Taken together, this suggests that understanding the cognitive processes that support rapid learning may require theorists to consider a broader space of planning strategies. 

Finally, our analysis of the RL strategies that emerge in transformers suggest a novel use of episodic memory-- each memory is not only a record of the original experience, but also stores additional computation useful for decision making \citep{dasgupta2021memory}.

% \section{Acknowledgements}
% We thank Ekdeep Lubana, Jack Lindsey, Gautam Reddy, Ben Eysenbach, Lily Zhang, and members of the Rajan lab for helpful discussions and feedback. This work was supported by the Harvard Medical School Lefler Small Grant Award and Dean’s Innovation  Award, NIH (RF1DA056403), James S. McDonnell Foundation (220020466), Simons Foundation (Pilot Extension-00003332-02), McKnight Endowment Fund, CIFAR Azrieli Global Scholar Program, and NSF (2046583).

\section*{Reproducibility}
Code will be publicly available in a Github link in the final paper (after de-anonymization).

\bibliography{iclr2026_conference}
\bibliographystyle{iclr2026_conference}

% \section{Appendix}

\newpage
\appendix

\section{Related Works}
\label{sec:appendix-related-work}

\textbf{Meta-learning to discover in-context reinforcement learning algorithms}\quad
In order to develop a model that can in-context reinforcement learn, we use a recently introduced meta-learning framework \citep{lee2023supervised}. Meta-learning is concerned with ``learning-to-learn'', using prior knowledge learned from previous tasks to support rapid adaptation to new ones \citep{beck2023survey}.
That is, the goal is to learn an algorithm $f$ that can be deployed in new tasks. The learning of $f$ is called the \textit{outer-loop} while $f$ itself is referred to as the \textit{inner-loop} \citep{beck2023survey}.
In many settings where $f$ is a RL algorithm (meta-RL), the outer loop shapes the weights of the network, but learning within each task occurs via activation dynamics-- through memory and internal state-- rather than parameter updates \citep{beck2023survey, lampinen2024broader}. The resultant $f$ is considered an in-context learning algorithm \citep{beck2023survey, sandbrink2024modelling, lampinen2024broader}.

Early examples of this approach include RL$^2$, which meta-trains a recurrent neural network using reinforcement learning in the outer loop, such that an in-context RL strategy emerges in the inner loop \citep{wang2016learning, duan2016rl}.
Subsequent work extended this approach to include explicit episodic memory mechanisms, combining RNNs with key–value memory architectures \citep{ritter2018been, ritter2020rapid, team2023human}.
More recently, \citet{lee2023supervised} proposed decision-pretrained transformers (DPTs), which use supervised training in the outer loop to induce in-context RL behavior in the inner loop.
We adopt DPTs in this work both for practical reasons--their scalability and ease of training--and for scientific ones: the transformer architecture allows us to probe how memory-based computation supports in-context learning.

\textbf{Meta-learning to describe cognition and neural activity}\quad Meta-learning has been proposed as a framework to model both cognitive flexibility and structured learning in neuroscience and psychology \citep{binz2024meta}.
In human cognitive tasks, meta-learned models have been used to replicate observed heuristics in decision-making and to account for few-shot generalization \citep{dasgupta2020theory, binz2022heuristics, lake2023human}.
Meta-RL, in particular, has been used to generate hypotheses about how neural systems implement learning across tasks. Previous studies have used the RL2 framework to show how the outer and inner learning loops can model different areas of the brain, with the prefrontal cortex often playing a key role \citep{wang2018prefrontal, hattori2023meta, zheng2025flexible}.
Despite these advances, many computational models in neuroscience rely on single-task RL, training agents independently on each task without leveraging prior experience.
In contrast, we use meta-RL as a tool for developing flexible in-context learning, without attempting to localize the outer loop to any specific brain region. Our focus is on the computational content of the learned representations and decision-making strategies.

\textbf{Transformers and episodic memory systems}\quad Transformer models process sequences by computing self-attention over key–value pairs, enabling flexible access to information across long contexts \citep{vaswani2017attention}.
This key–value structure has led to interpretations of transformers as memory systems \citep{geva2020transformer}, aligning them with a broader class of models that incorporate explicit memory mechanisms \citep{graves2014neural, sukhbaatar2015end, graves2016hybrid, banino2020memo}.
These systems separate memory addressing (keys) from memory content (values), enabling high-fidelity storage and targeted retrieval.
This architectural separation bears many similarities to theoretical accounts of episodic memory in the brain \citep{teyler1986hippocampal, teyler2007hippocampal, gershman2025key}.
For example, recent work has formalized connections between key–value architectures and Hopfield networks, a classic model of associative memory in the brain \citep{krotov2020large, ramsauer2020hopfield}.
Related approaches such as fast-weight models \citep{ba2016using, munkhdalai2017meta} offer alternative mechanisms for temporary memory storage and in-context computation, often drawing from Hebbian or synaptic dynamics.
Other studies have proposed biologically grounded implementations of key–value attention mechanisms, further linking transformer-like architectures to neural computation \citep{bricken2021attention, tyulmankov2021biological, whittington2021relating, kozachkov2023building, fang_2025, chandra2025episodic}.
Several of these models take direct inspiration from the hippocampus, a brain region widely implicated in episodic memory \citep{whittington2021relating, fang_2025, chandra2025episodic}.
Recent experimental work has also identified key–value–like coding patterns in hippocampal activity during episodic memory tasks \citep{chettih2024barcoding}.
Together, these findings suggest that key–value architectures offer a useful computational analogy for episodic memory.
In this work, we analyze a meta-trained transformer to examine what kinds of memory-supported strategies emerge when such an architecture is optimized for rapid adaptation.

\section{Task construction}
\label{sec:appendix-task-construction}
\subsection{Gridworld}
We use a $5 \times 5$ 2D gridworld environment. Thus, there are $N=5\times 5=25$ states in the environment, each of which corresponds to an underlying $(x,y)$ location. Actions are one-hot encoded and consist of: up, right, down, left, stay. If the agent chooses to take an action that hits the environment boundaries, this manifests as a "stay" transition. The transition structure in this environment $\mathcal{T}$ (that is, how actions transition the agent from one $(x,y)$ state to another) is fixed across all tasks. Each task is defined by the sensory encoding $\mathcal{S}$, the reward location $\mathcal{R}$, and the in-context exploration trajectory $\mathcal{D}$.

Each state in a gridworld task corresponds is encoded by a $10$-dimensional vector. For each task, we describe the set of these $N$ encoding vectors as $\mathcal{S}$. The following describes how we generate the encoding vectors comprising $\mathcal{S}$. We first define a random expansion matrix $E\in \mathcal{R}^{N\times N}$, where $E_{i,j} \in \mathcal{N}(0,1)$. We next construct a distance correlation matrix $D\in \mathcal{R}^{N\times N}$ by exponentiating the negative Euclidean distances between all pairs of grid positions: $D_{i,j} = \sigma^{(||(x_i, y_i) - (x_j, y_j)||_2)}$ for states $i$ and $j$ and their corresponding $(x,y)$ locations. Here, $\sigma\in [0,1]$ is a correlation parameter that controls how strongly nearby positions are correlated in the encoding space. Thus, the encoding of state $i$ is computed as $\frac{ED_{:,i}}{||ED_{:,i}||_2}$ and $\mathcal{S}=\{\frac{ED_{:,i}}{||ED_{:,i}||_2}\}_{i=0}^N$. 

The reward state $\mathcal{R}$ is chosen from the $N$ states in the environment. The in-context exploration trajectory is generated from a random walk with a randomly chosen initial state, plus some reasonable heuristics. Specifically, we make the probability of selecting the "stay" action half as likely as the other actions. In addition, if the agent takes an action that causes it to not transition to a new state, the probability of taking that action again is downweighted to $0$ until the agent transitions to a new state (preventing the agent from getting stuck at boundaries). Running this biased random walk for $T$ steps gives us $C={(s_t, a, s'_t, r_t)}_{t=0}^T$, a set of standard RL transitions. 

Our dataset is generated offline before training. We now describe how we construct the train/evaluation/test sets. For a desired dataset size of $M$, we partition the $N$ states of the gridworld environment into three sets of sizes $M*p_{train}$, $M*p_{eval}$, $M*p_{test}$. Specifically, we divide the dataset with ratios: $p_{train}=0.8$, $p_{eval}=0.1$, $p_{test}=0.1$. To generate the training dataset, we construct $M*p_{train}$ tasks, where we sample $\mathcal{R}$ from the corresponding training partition of states. We then sample $\mathcal{S}$ and $\mathcal{D}$ as described above. We repeat this for the evaluation and test datasets. The training dataset is used for pretraining. The evaluation dataset is used for validation during pretraining and selecting models. The test dataset is used for any analyses conducted after model training and selection.

\subsection{Tree Mazes}
We use binary tree environments with $7$ tree layers (that is, a minimum of $6$ actions is needed to navigate from root to leaf). Actions are one-hot encoded and consist of: right child, left child, parent, stay. If the agent tries to transition to a node that does not exist (e.g. trying to go to "parent" from the root), this manifests as a "stay" transition. The transition structure $\mathcal{T}$ can vary across tasks. This is because in each task the underlying tree is generated with branching probability $0.9$. Thus, there is a maximum of $127$ states in each task. Each task is defined by the transition structure $\mathcal{T}$, sensory encoding $\mathcal{S}$, the reward location $\mathcal{R}$, and the in-context exploration dataset $\mathcal{D}$.

The sensory encodings are generated as in gridworld, except $D$ is defined via the geodesic distances between any two tree states. The reward state $\mathcal{R}$ is chosen only from leaf nodes. The in-context exploration trajectory $\mathcal{D}$ is generated from a random walk from the root node, with reasonable heuristics. We use the same heuristics as in gridworld. We also add heuristics described in \citet{rosenberg2021mice} of mice in similar mazes. That is, the agent is more likely to alternate between left and right transitions when transitioning through child nodes. In addition, the agent is less likely to backtrack. 

As before, our dataset is generated offline before training. For a desired dataset size of $M$, we construct three sets of sizes $M*p_{train}$, $M*p_{eval}$, $M*p_{test}$, with $p_{train}=0.8$, $p_{eval}=0.1$, $p_{test}=0.1$. To generate the training dataset, we construct $M*p_{train}$ tasks, where we sample $\mathcal{T}$ from binary trees with branching probability $0.9$. We use only trees with at least one leaf node in the seventh layer and, for the training dataset, exclude the full $7$-layer tree. We sample $\mathcal{R}$ from one of the leaf nodes. We then sample $\mathcal{S}$ and $\mathcal{D}$ as described above. We repeat this for the evaluation dataset, ensuring distinct $\mathcal{T}$ from the training dataset. The test dataset comprises only of $\mathcal{T}$ corresponding to a full binary tree, with $\mathcal{R}, \mathcal{S}, \mathcal{S}$ sampled as above. As b before, the training dataset is used for pretraining. The evaluation dataset is used for validation during pretraining and selecting models. The test dataset is used for any analyses conducted after model training and selection.

\section{Model and Training Parameters}
\label{sec:appendix-model-training}
We largely follow the same architecture as that of \citet{lee2023supervised}, a GPT-2 style model with causal attention and without positional embeddings. Our default model has 4 heads, 3 layers, and embedding dimension of $512$. Context memory tokens consist of the $(s_t, a, s'_t, r_t)$ tuple concatenated together into one vector. Thus, tokens are $26$-dimensional in gridworld and $25$-dimensional in tree maze. The query token consists of $(s_q, \vec{0})$ for query state $s_q$, where $\vec{0}$ provides 0-padding to reach the desired vector size. These tokens are projected into model embedding space through a learnable linear layer. The model samples greedily in the gridworld environment and with softmax sampling in the tree maze environment (both settings were empirically determined).

In contrast to \citet{lee2023supervised}, we provide the query token at the end of the context memory. This is to allow a clearer interpretation in which context memory tokens represent previous experiences in the maze that are stored in episodic memory. The query token is only provided at decision time, and the agent must use previous memories to guide its present decision.

To allow for query tokens at the end of an input sequence sequence and to preserve efficient pretraining, we make modifications to the pretraining procedure of \citet{lee2023supervised}, which we describe here. Let's say we have a pretraining task with context memory tokens $\mathcal{D}$ and query state $s_q$. To encourage length generalization, we would like to train the model on many sequence lengths-- let's say every $t_{step}$ timesteps of $\mathcal{D}$. To do so from one forward pass, we first construct a sequence $\mathcal{D}_{train}$ where $s_q$ is interleaved every $t_{step}$ timesteps of $\mathcal{D}$: $\mathcal{D}_{train}=[\mathcal{D}_1,\mathcal{D}_2, \dots, \mathcal{D}_{t_{step}}, s_q, \mathcal{D}_{t_{step}+1},\dots  \mathcal{D}_{2*t_{step}}, s_q, \mathcal{D}_{2*t_{step}+1},\dots, \mathcal{D}_{T}, s_q]$. We then construct an attention mask $A_{mask}=A_{causal}+A_{query}$, where $A_{causal}$ is the standard causal attention mask with $-\infty$ values in the upper-triangular and $0$ elsewhere. $A_{query}$ ensures that query tokens are not processed by context memory tokens by masking columns corresponding to the query token:
\begin{equation}
    A_{query}[i,j]=
    \begin{cases}
      -\infty, & \text{if}\ \mathcal{D}_{train}[j]=s_q \text{ and } i \neq j \\
      0, & \text{otherwise}
    \end{cases}
\end{equation}
Thus, we use $A_{mask}$ during training and cross-entropy loss is only calculated over the outputs corresponding to $s_{q}$. In gridworld, the maximum context length in training is $T=200$. In tree maze, the maximum context length in training is $T=800$.

We use Adam optimizer with weight decay $1\times10^{-5}$. We use a batch size of $1024$ for gridworld and a batch size of $512$ for tree mazes. We train the model for 25 epochs in gridworld and 50 epochs in tree mazes. We use a learning rate of $1\times10^{-4}$ which we linearly decrease to $1\times10^{-5}$ over the course of training (we found that this empirically worked well). For each training run, we use two NVIDIA H100 GPUs. This results in around $1$ hour of training time for gridworld and $1.5$ hours of training time for tree mazes. For each task, we train models from $5$ random initializations, for both a dropout of $0$ and $0.2$. We then select the best model via validation performance for each task. This selected model is the model we analyze for each task.

\begin{figure}[t!]
  \centering
  \includegraphics[width=\textwidth]{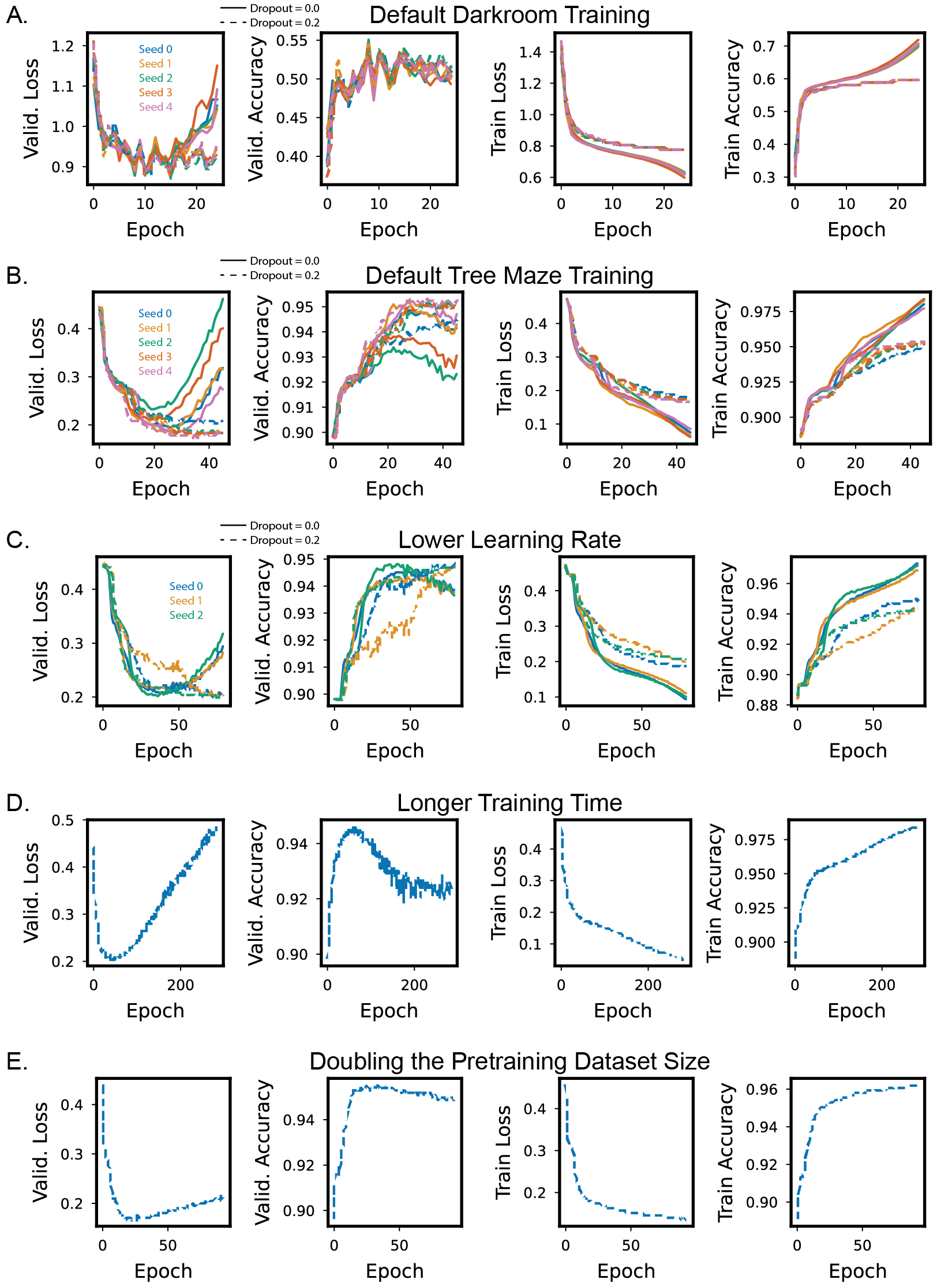}
  \caption{Effect of training parameters. \textbf{A.} Default training settings for Darkroom task, showing validation loss, validation accuracy, training loss, and training accuracy over training epochs. Colors indicate random seeds and line style indicates dropout amount. We note that $100\%$ accuracy is not possible due to the training procedure (see task construction details). \textbf{B.} As in (A), but for tree mazes. \textbf{C.} As in (B), but for $\frac{1}{10}$ of the default learning rate. We don't use a learning rate scheduler here. \textbf{D.} Seed 0 of (B), but we let the training run for 250 epochs. \textbf{E.} As in (B) but pretraining dataset is doubled in size.}
  \label{fig:8}
\end{figure}

\begin{figure}[t!]
  \centering
  \includegraphics[width=\textwidth]{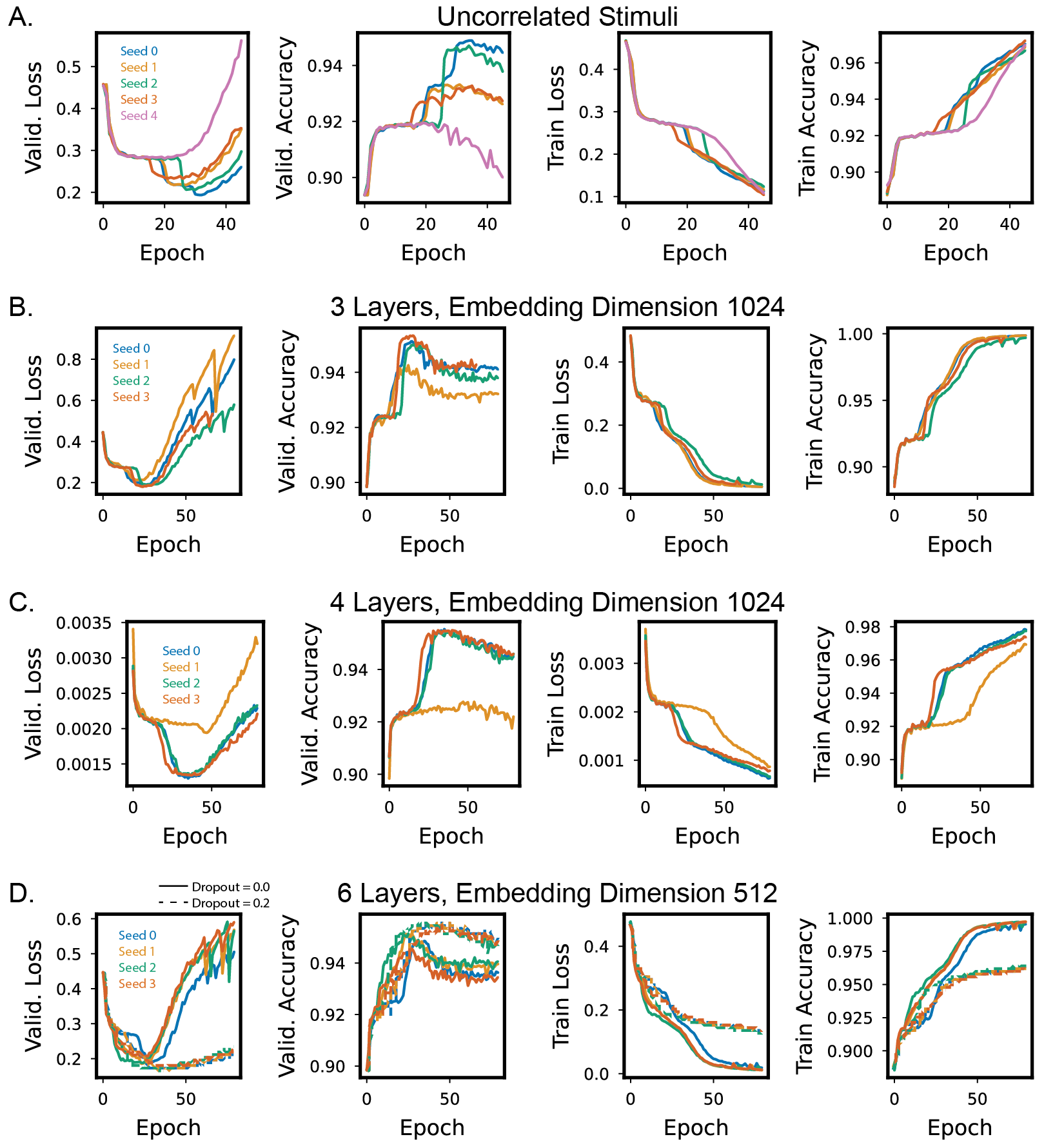}
  \caption{Effect of model and dataset parameters on training. \textbf{A.} As in Fig \ref{fig:8}B, but state encoding is fully uncorrelated. \textbf{B.} As in Fig \ref{fig:8}B, but model encoding dimension is doubled to 1024. \textbf{C.} As in Fig \ref{fig:8}B, but model encoding dimension is doubled to 1024, and number of model layers is increased to 4. \textbf{D.} As in Fig \ref{fig:8}B, but number of model layers is doubled to 6.}
  \label{fig:9}
\end{figure}

We did not find much improvement by trying other tweaks to the pretraining setup or model size. We show validation and training results for various different parameters in figure \ref{fig:8} and \ref{fig:9}.

\newpage
\section{Results Sensitivity to Task/Model Parameters}
\label{sec:appendix-results-sensitivity}

We repeat some of the analyses in the main figures here for alternative parameterizations of task and model. We focus mostly on testing the tree maze environment, for simplicity. 

\begin{figure}[t!]
  \centering
  \includegraphics[width=\textwidth]{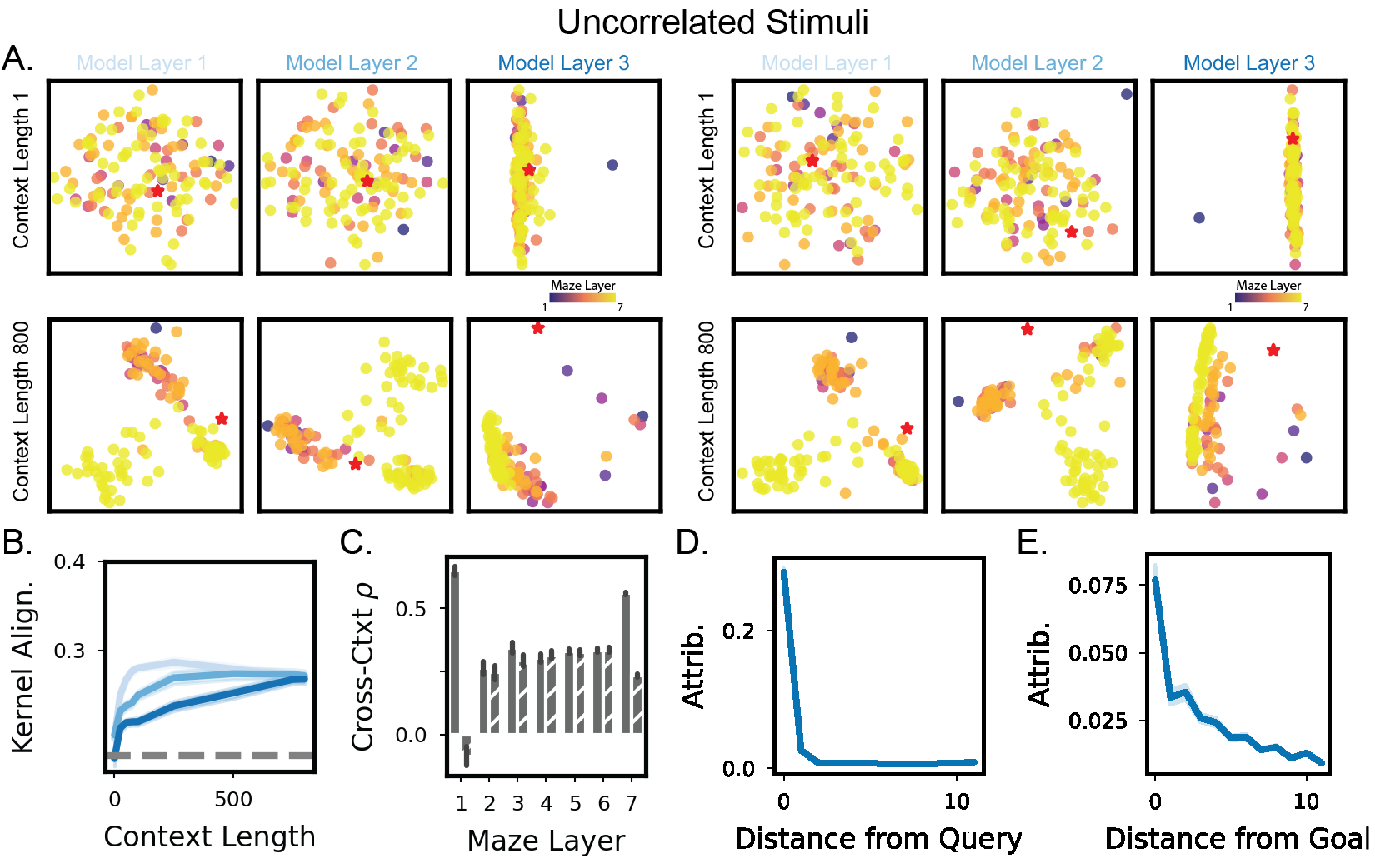}
  \caption{Sensitivity of results to state encoding correlation. \textbf{A.} As in Fig \ref{fig:3}E, but for two random example environments (left and right). Additionally, the model was pretrained and tested on environments with uncorrelated stimuli. \textbf{B.} As in Fig \ref{fig:3}C, but for uncorrelated stimuli. \textbf{C.} As in Fig \ref{fig:4}F, but for uncorrelated stimuli. \textbf{D., E} As in Fig \ref{fig:5}BC, but for uncorrelated stimuli.}
  \label{fig:10}
\end{figure}

We first test the model where stimuli do not have spatial correlation (Fig \ref{fig:10}). We find similar coarse in-context representation structure emerges, where representations from the first layer roughly separate out the two main branches of the maze (Fig \ref{fig:10}AB). However, the bifurcating structure is less clear than it is when some spatial correlation is introduced (Fig \ref{fig:3} and Fig\ref{fig:16}). The cross-context structure results seem similar to that of the correlated stimuli (Fig \ref{fig:10}C, compare to (Fig \ref{fig:4}F). The bias of attribution scores to the query and goal at decision time also mirrors that seen in environments with correlated stimuli (Fig \ref{fig:10}DE, compare to (Fig \ref{fig:5}BC). 

\begin{figure}[h]
  \centering
  \includegraphics[width=\textwidth]{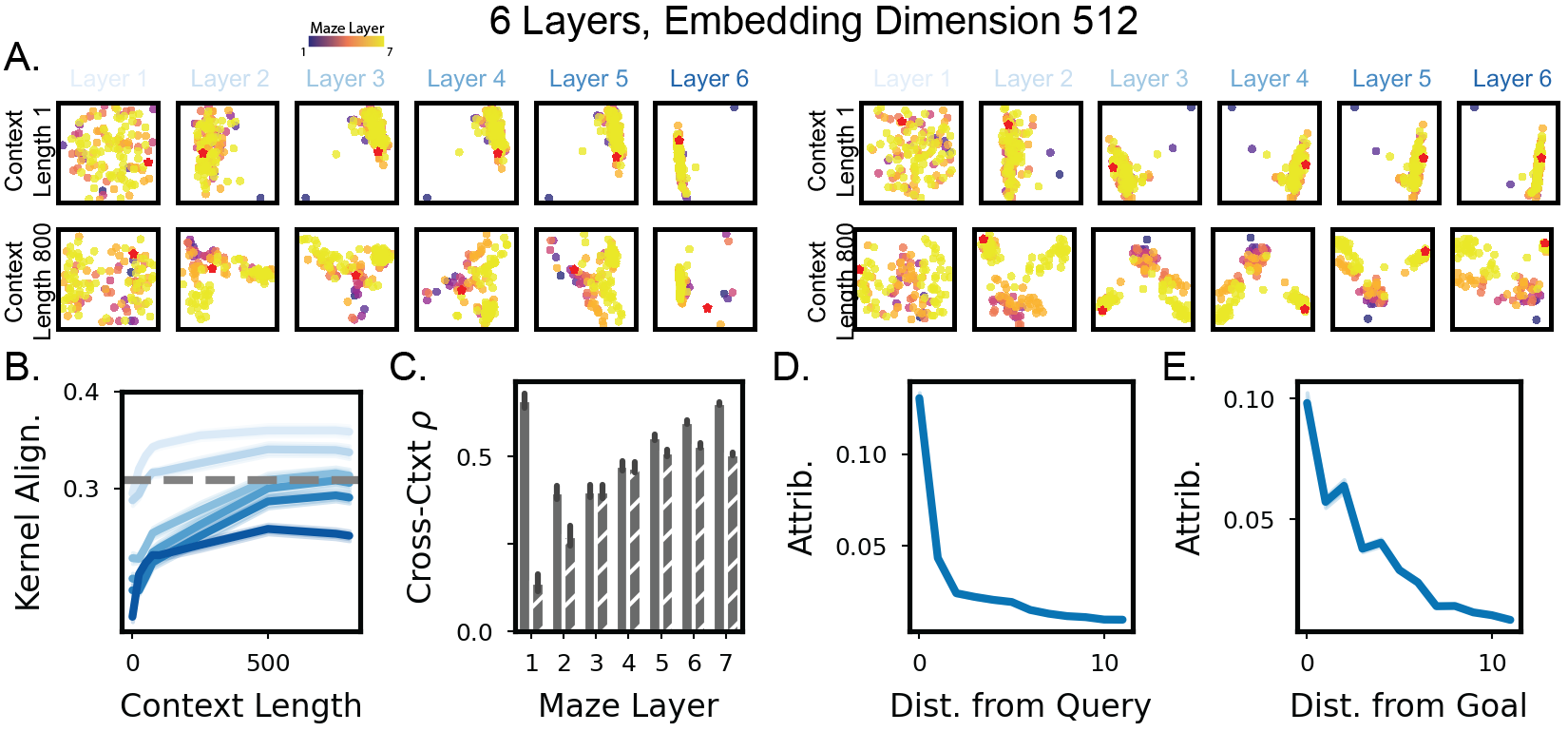}
  \caption{Sensitivity of results to model size. \textbf{A.} As in Fig \ref{fig:3}E, but for two random example environments (left and right). Additionally, the model has twice the number of layers (6). \textbf{B.} As in Fig \ref{fig:3}C, but for 6-layer model. \textbf{C.} As in Fig \ref{fig:4}F, but for 6-layer model. \textbf{D., E} As in Fig \ref{fig:5}BC, but for 6-layer model.}
  \label{fig:11}
\end{figure}

We next test a larger version of the model with 6 layers (Fig \ref{fig:11}). Like before, the in-context representation structure emerges as a bifurcating structure in the middle layers of the model (Fig \ref{fig:11}AB). With more layers, though, the representations of the first layer now appear disorganized. As before, the last layer of the model is also organized in a less interpretable structure. The cross-context structure results again reflect greater latent structure alignment in the early and late layers of the model (Fig \ref{fig:11}C, compare to (Fig \ref{fig:4}F). The bias of attribution scores to the query and goal at decision time also mirrors that seen in environments with correlated stimuli (Fig \ref{fig:11}DE, compare to (Fig \ref{fig:5}BC). 

\newpage
\section{Additional Behavioral Results in Gridworld and Tree Mazes}
\label{sec:appendix-fig_2}
\begin{figure}[h!]
  \centering
  \includegraphics[width=\textwidth]{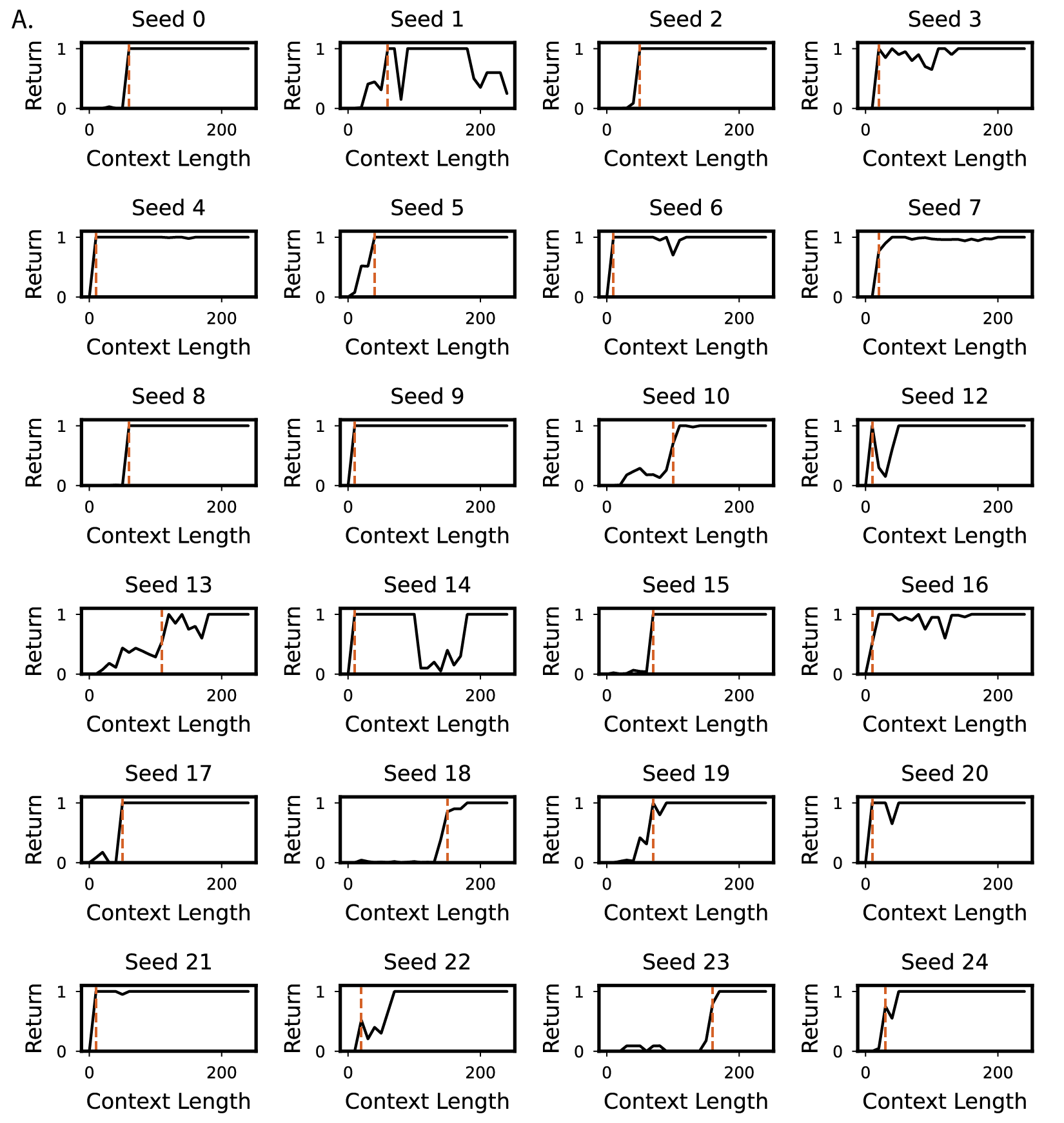}
  \caption{Additional in-context learning curves for gridworld task. \textbf{A.} As in Fig \ref{fig:2}A, but for 24 additional test environments. We skipped environments where reward was never seen in-context.}
  \label{fig:12}
\end{figure}

Here, we show additional learning results for both the gridworld task and tree mazes. We first show additional in-context learning curves for gridworld (Fig \ref{fig:12}). As in the main results, performance is evaluated from query states that were seen in-context and at least 6 steps away from the goal. If no eligible query states meet the selection criteria, return is recorded as zero. We note that in-context learning can still be unstable at times. In part, this may be because the model is sometimes tested on states it has not experienced. Thus, it is more difficult to navigate into previously experienced territory to find the goal. We also suspect that improvements in the training procedure or architecture that we have not explored could also produce a more performant model.

\begin{figure}[h!]
  \centering
  \includegraphics[width=\textwidth]{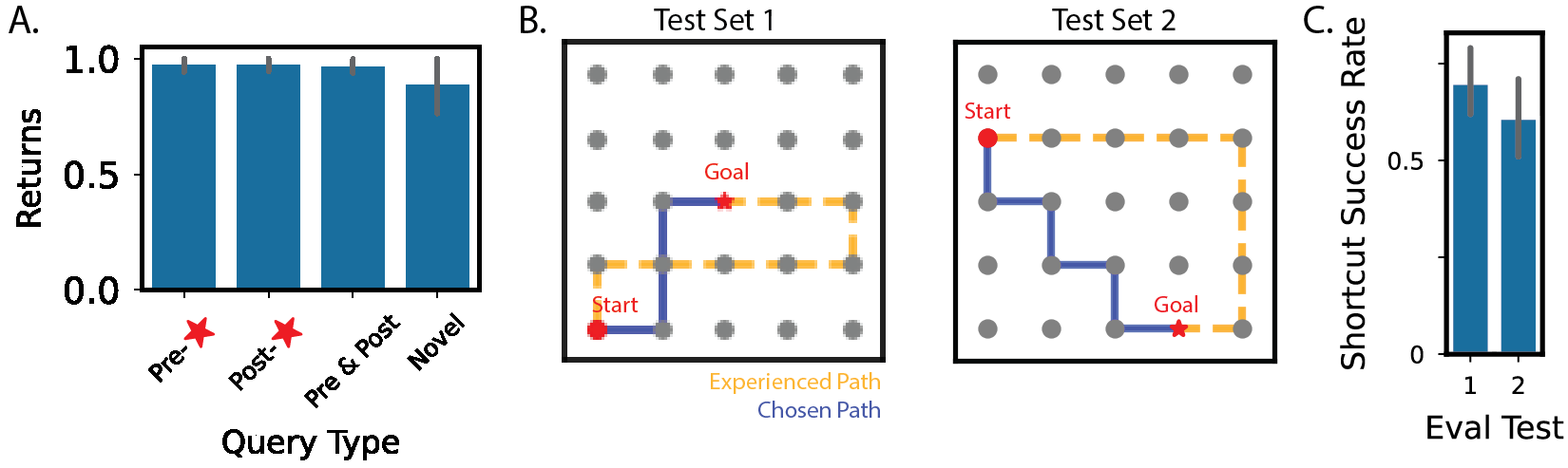}
  \caption{Additional learning results in gridworld task. \textbf{A.} As in Fig \ref{fig:2}F, but for gridworld task. We also further divide the query states into states seen only before any reward experience (Pre-\faStarO), states seen only after all reward experiences (Post-\faStarO), states seen before and after reward experiences (Pre \& Post), and states that were never seen in-context (novel). \textbf{B.} Depictions of the two tests for shortcut paths we use (left and right). Each test set has a fixed start location, goal location, and in-context experienced path (yellow dashed line). For each test set, we simulate 100 environments with different sensory encodings. Blue line shows an example successful shortcut path taken by the model. See methods description for more details. \textbf{C.} Success rate of taking the optimal, shortcut path for the two test sets in (B), across 100 sample environments. Error bars show $95\%$ confidence interval. Note that chance level in both tests is $0.02$ (generously excluding the stay transition from consideration).} 
  \label{fig:13}
\end{figure}

A few more behavioral results are shown for gridworld. We reproduce the analysis of Fig \ref{fig:2}F for the meta-learned gridworld model (Fig \ref{fig:13}A). We further subdivide the query states, however. This is because we were curious if the agent would perform differently for states seen only before any reward experience or states seen only after all reward experience. This turns out not to be the case, and the agent does equally well in both cases (Fig \ref{fig:13}A). This information is useful for forming hypotheses of how the model solves the task. Due to its causal structure, this means that the model probably doesn't (solely) rely on a strategy where experiences of reward alter the processing of subsequent context memory tokens. Otherwise, the model should do poorly on states that were only seen before any reward experiences. We also give further details of the shortcut paths experiments (Fig \ref{fig:13}BC).

\begin{figure}[h!]
  \centering
  \includegraphics[width=\textwidth]{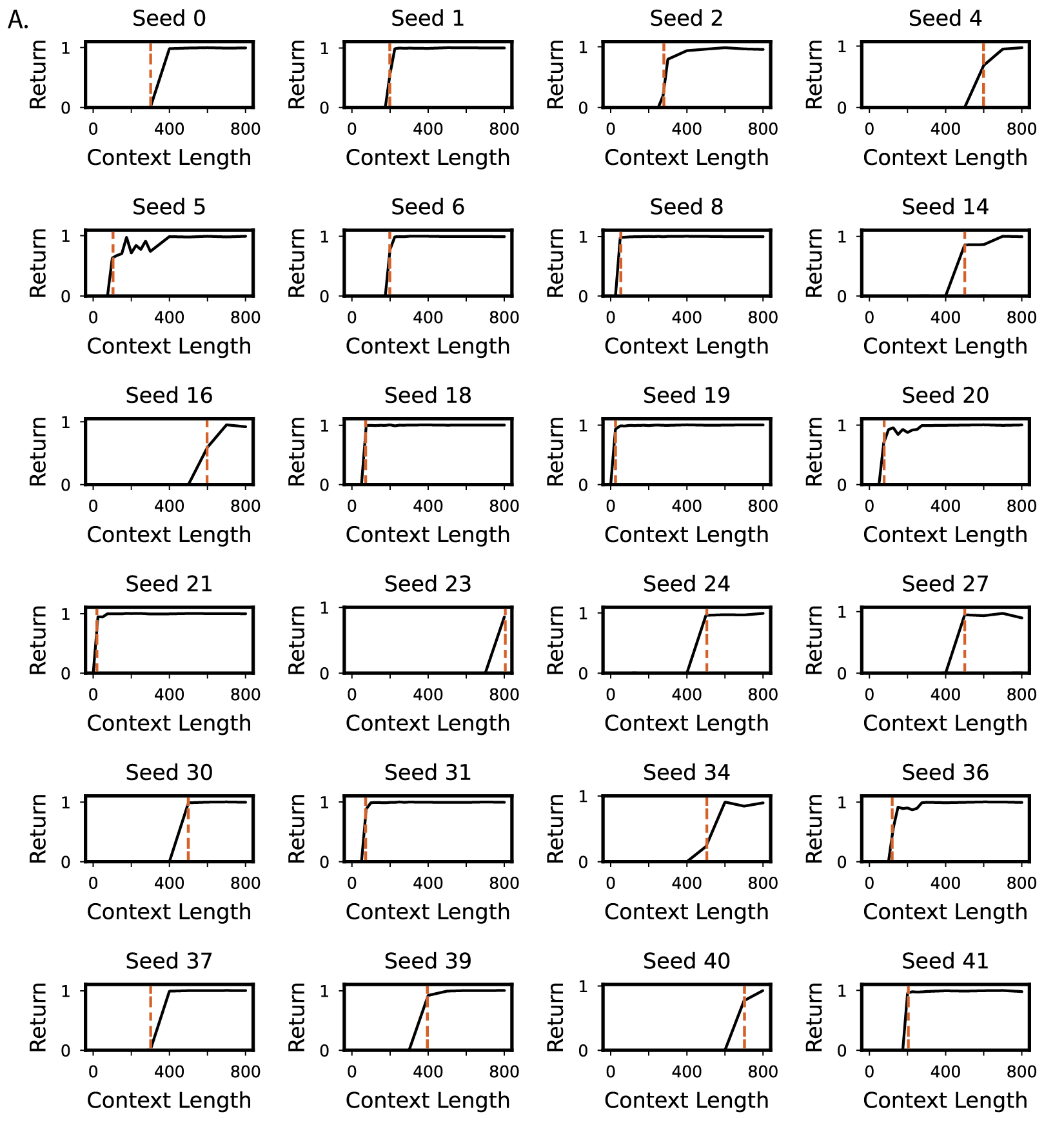}
  \caption{Additional in-context learning curves for tree maze task. \textbf{A.} As in Fig \ref{fig:2}D, but for 24 additional test environments. We skipped environments where reward was never seen in-context.}
  \label{fig:14}
\end{figure}

Finally, we show additional in-context learning examples for tree mazes (Fig \ref{fig:14}). Learning is much more stable in this environment, perhaps because there are useful heuristics the agent can use if at a novel state (transition towards parent node until arriving at a state that has already been experienced).

\newpage

\newpage

\newpage

\newpage
\section{Q-Learning Simulations}
\label{sec:appendix-q-learning}
To make comparisons to RL algorithms without learned priors, we simulate two Q-learning models. We use a tabular Q-learning model where we abstract away the problem of representation learning and allow the model to use a lookup table. We also use a deep Q network (DQN) parameterized as a MLP with 4 hidden layers of dimensions $[256, 128, 64, 16]$.

To make as fair a comparison as possible, we give our Q-learners a full replay buffer and let models train to convergence on the memories of the buffer. For instance, given a task with context memory $\mathcal{D}$ if we wish to evaluate the model at context length $t_C$ we define a replay buffer comprising $\mathcal{D}_{1:t_C}$. We then let the Q-learning model train on several epochs over the full dataset of the replay buffer, until the temporal difference error has converged. We find that $1000$ epochs for the tabular model and $1500$ epochs for the DQN is more than sufficient to ensure this. For the tabular model, we train with batch size 512 and learning rate $0.1$. For the DQN, we train with batch size 1024 and learning rate $1\times10^{-5}$. 

There are also a few additional training details for the DQN. We randomly reinitialize the network weights at each context length before we run the training procedure. This is because we find that resetting the weights works empirically better than initializing with the weights from the previous context length the model was trained on (this is reasonable, as the latter induces a continual learning problem). To maintain as many parallels to standard methods as possible, we also adopt a double deep Q learning framework \citep{van2016deep}. We use a target network that is updated every $10$ epochs. We don't think this detail is critical (and empirically the use of a target network here doesn't seem to impact performance) as the learning problem in our setting is fully stationary.

For both models, at test time we also allow for action sampling with some temperature. We empirically select the temperature that results in the best performance after a grid search over the values $[0.005, 0.01, 0.05, 0.1, 0.2, 0.5, 10.0]$. We also selected a value of $\gamma$ in the TD loss function that worked well in practice after a grid search over the values $[0.7, 0.8, 0.9]$: $\gamma=0.8$ for the tabular model, $\gamma=0.9$ for DQN. 

\newpage
\section{Additional in-context representation learning results}
\label{sec:appendix-in-context}
\begin{figure}[h!]
  \centering
  \includegraphics[width=\textwidth]{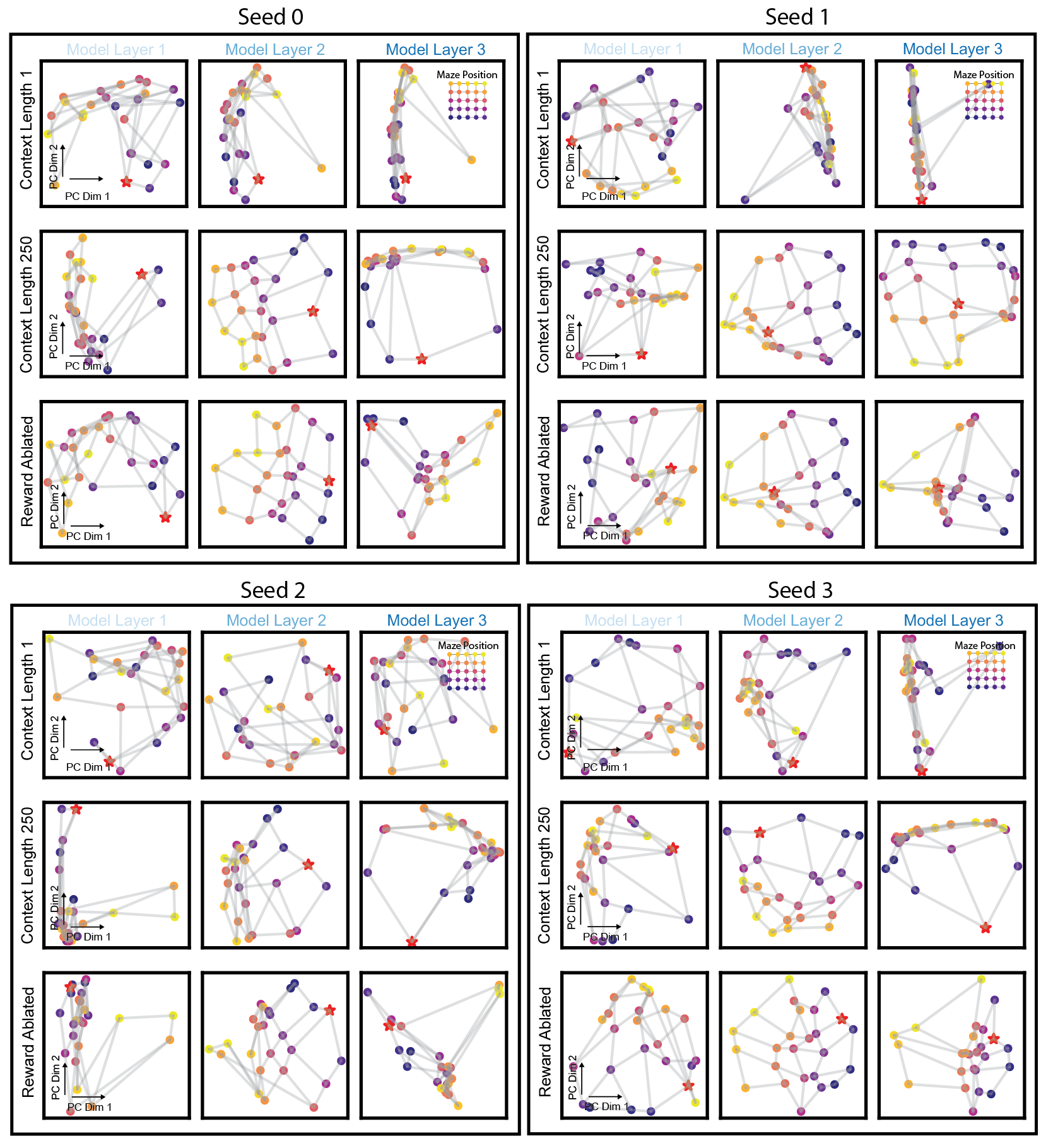}
  \caption{Additional in-context representation learning examples in gridworld task. As in Fig \ref{fig:3}A, but for four more additional random seeds. Additionally, the third row of each plot shows the PCA embedding plots at context length 250 if reward was ablated ($r=0$ in all transitions).}
  \label{fig:15}
\end{figure}

\begin{figure}[h!]
  \centering
  \includegraphics[width=\textwidth]{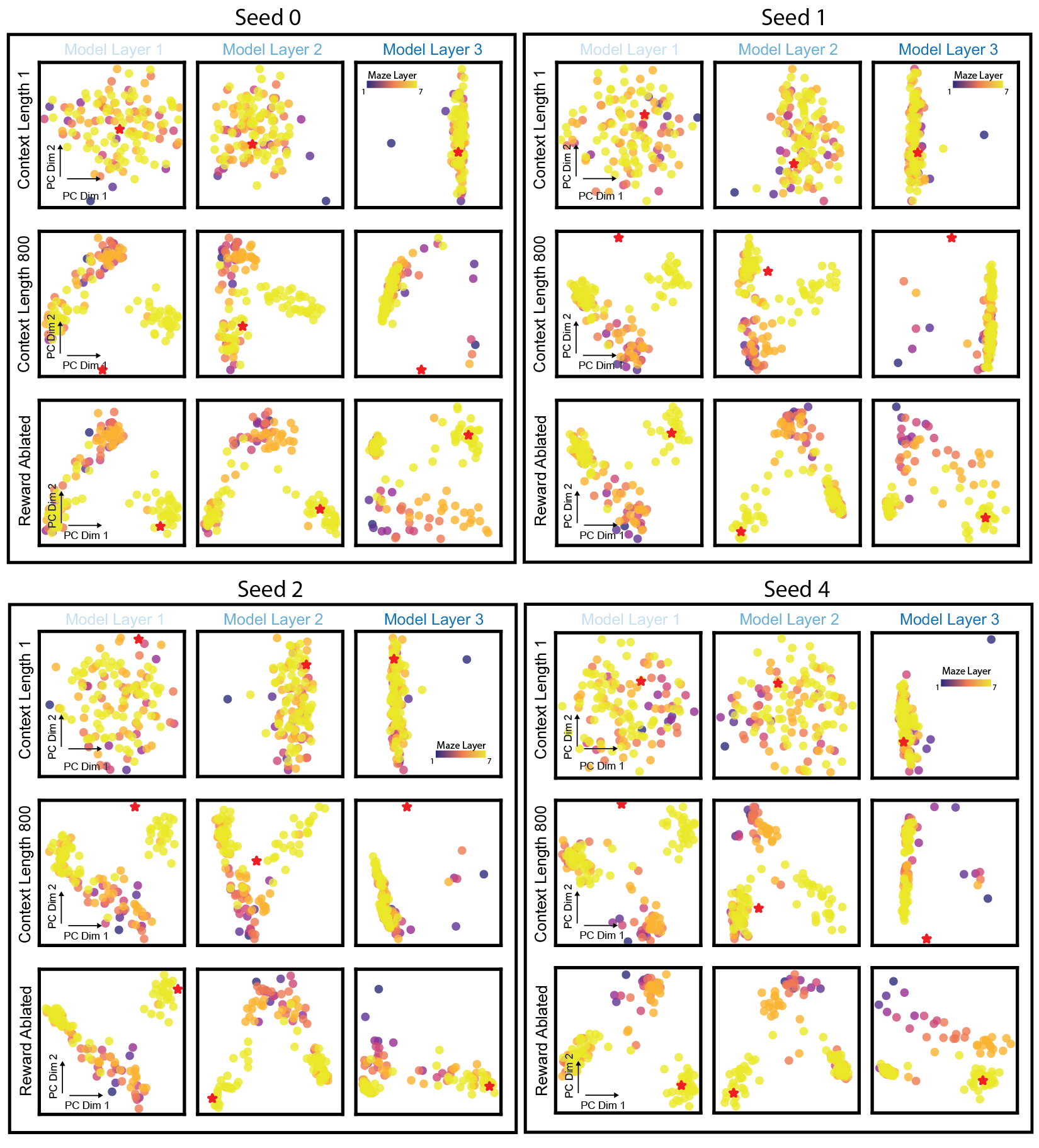}
  \caption{As in Fig \ref{fig:3}E, but for four more additional random seeds. Additionally, the third row of each plot shows the PCA embedding plots at context length 800 if reward was ablated ($r=0$ in all transitions). We skipped seeds where rewards was already never seen during in-context exploration.}
  \label{fig:16}
\end{figure}

Here, we show additional in-context representation learning examples for more randomly sampled environments (Fig \ref{fig:15} for gridworld, Fig \ref{fig:16} for tree maze). In addition, we show the results of reward ablation on representation learning. Comparing the model with and without reward ablation, it appears that reward information sometimes results in the reward state being pushed farther away from non-rewarding states.

\newpage

\newpage
\section{Kernel Alignment}
\label{sec:appendix-kernel-alignment}
To quantify how well learned model representations capture the structure of the latent environment, we employed centered kernel alignment (CKA) to compare the similarity between the true environment structure and the model's internal representations. We first constructed a ground truth kernel matrix by computing the environment distance matrix $D$ where $D_{ij}$ indicates the number of actions needed to navigate from state $i$ to state $j$. We then applied an exponential transformation $K_{\text{input}} = \gamma^D$ where $D$ is the distance matrix and $\gamma$ controls the spatial scale of environment structure captured by the kernel.

For each network layer, we extracted hidden state representations corresponding to each environment state, using the final token representation as the state embedding. We collect this in the matrix $X$ and construct representation kernel $K_{\text{latents}} = (X - \bar{X})(X - \bar{X})^T$. We then compute the CKA between $K_{\text{input}}$ and $K_{\text{latents}}$.

\begin{figure}[h!]
  \centering
  \includegraphics[width=\textwidth]{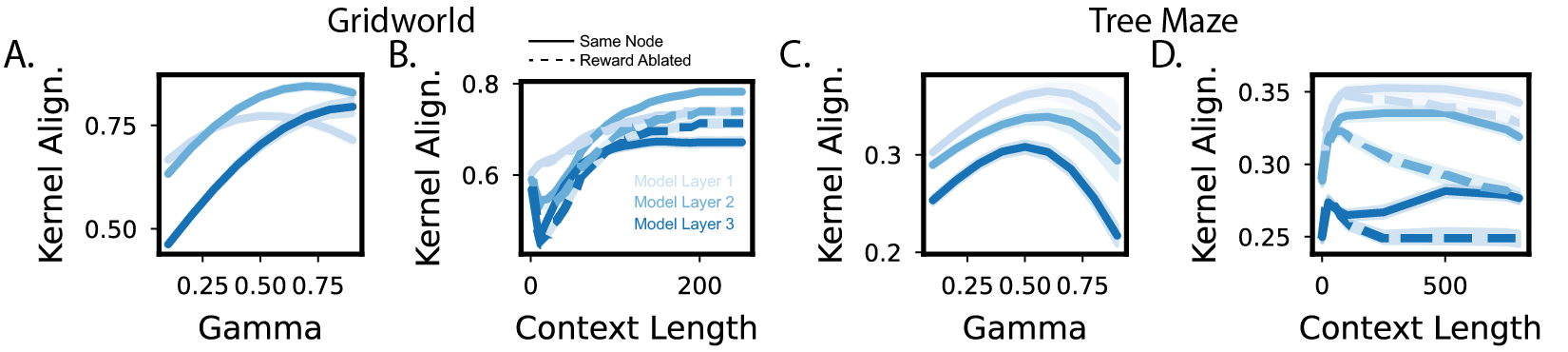}
  \caption{Additional kernel alignment details. \textbf{A.} Kernel alignment for different values of $\gamma$, the spatial kernel used to define the environment structure. Line colors indicate model layer where representations are extracted. \textbf{B.} As in Fig \ref{fig:3}C, but showing additional lines (dashed) where reward was ablated ($r=0$ in all transitions). \textbf{C, D} As in (A, B), but for the tree maze task.}
  \label{fig:17}
\end{figure}

In Fig \ref{fig:17}AC, we show how the kernel alignment score changes for different values of $\gamma$. For the analyses in the main text, we select a value of $\gamma$ that maximizes overall kernel alignment: $0.8$ for gridworld, and $0.6$ for tree mazes.

We note that the kernel alignment measure is likely still imperfect for what we want to quantify, especially in tree mazes. For instance, the bifurcating structure of representations in tree mazes is an interesting phenomena we would like to understand better. However, it is a coarse structure that likely does not align well to the ground truth kernel that we defined. For instance, in Fig \ref{fig:17}BD, we see how reward ablations affect kernel alignment. In the tree maze task, there appears to be a significant difference in kernel alignment that is induced by reward ablations. In contrast, the PCA plots from Fig \ref{fig:16} show that the branching structure of the representations is well-preserved even when rewards are ablated. Thus, we think additional metrics may be more useful to interpret representation organization in tree mazes.

\newpage
\section{Analyzing Cross-Context Learning in Gridworld}
\label{sec:appendix-cross-context}
\begin{figure}[h!]
  \centering
  \includegraphics[width=\textwidth]{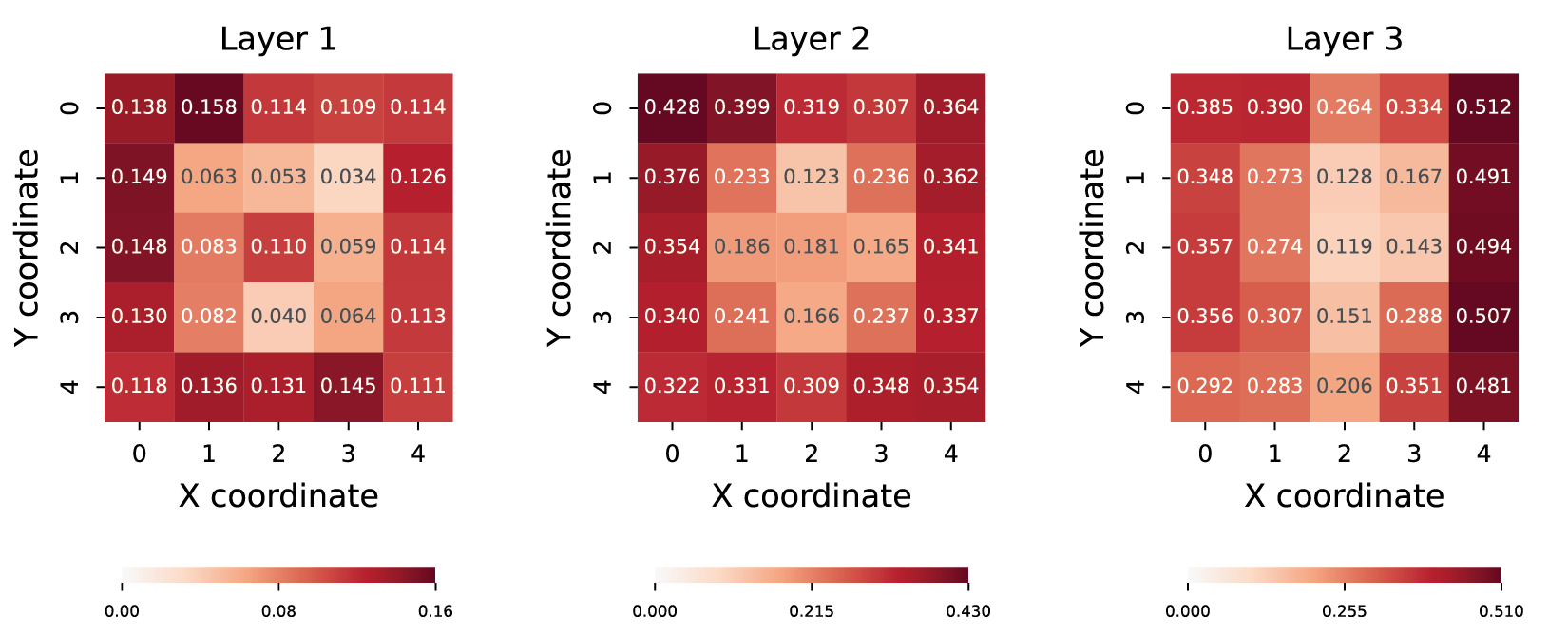}
  \caption{Additional results for cross-context representations in the gridworld task, for each of the three layers of the model. Each plot shows the difference between same-node and different-node correlations at context length 250 (that is, the difference between solid and dashed lines in Fig \ref{fig:4}B). Values are separated by the same-node identity, i.e. the underlying XY latent state.}
  \label{fig:18}
\end{figure}

Here, we show additional results for cross-context representation learning in gridworld. We plot the difference between same-node and different-node correlations and separate these values by the underlying XY latent state. Potentially, representations are better aligned across contexts at the edges of the environment (see ``Layer 1'' and ``Layer 2'' of Fig \ref{fig:18}). Overall, though, the cross-context similarity is fairly similar across the entire gridworld structure.

\newpage
\section{Linear Decoder Setup}
\label{sec:appendix-linear-decoder}
We will first describe how we linearly probe the representations of query state tokens. In gridworld, we randomly select $600$ tasks from the original train set and partition these tasks into a new train/test set for our linear decoder, with a $90/10$ split. In gridworld, we use the original train set because there's more unique XY goal locations (21) than in the original eval and test sets (2 each). In tree maze, we randomly select $600$ from the original test set and make the same $90/10$ train/test split for our linear decoder. We skip over tasks where reward is never seen during the in-context exploration phase.

The regressors for our decoder will be model representations at some layer. To collect them, in each environment we first identify the set of states that had been seen in-context. For each state $s'$ that was seen, we let query state $s_q=s$ and present the in-context exploration trajectory and $s_q$ as inputs to the model. For each model layer $l$, we collect the model representations for the $s_q$ token, $r(s_q, l) \in R^{512}$. The decoding task is to predict some value $v$ given $r(s_q, l)$, where $v$ is typically some kind of information pertaining to $s_q$. We tried a variety of values $v$ and in the main text only discuss the variables for which test decoding accuracy was high. 

To fit a linear decoder, we use ridge regression. We standardize features to $0$ mean and unit variance. The regularization strength $\alpha$ was selected through 5-fold cross-validation using a grid search over regularization strengths from $[10^0, 10^4]$, with 10 logarithmically-spaced values. Cross-validation was performed with shuffled splits. For each $\alpha$, we computed the mean $R^2$ score across all validation folds and selected the $\alpha$ that maximized this cross-validation performance. The final decoder for each layer was fit on the complete training set using the $\alpha$ found previously. Model performance was evaluated on the held-out test set.

For circular variables such as angles, we cannot directly apply standard regression since the circular nature of the data violates the assumptions of linear models (e.g., an angle of $\pi$ and $-\pi$ represent the same direction but appear numerically distant). Instead, we decompose each target angle $\theta$ into its sine and cosine components: $\text{sin}(\theta)$ and $\text{cos}(\theta)$. We then fit two separate ridge regressors to predict these components independently, using the same cross-validation procedure described above. To obtain the final angle prediction, we convert the predicted sine and cosine values back to angles using the arctangent function: $\hat{\theta}=\text{arctan2}(sin(\hat{\theta}), cos(\hat{\theta}))$.

For classification tasks, we do the same but with logistic regression and report balanced accuracy scores.

To probe representations for context memory tokens, we follow a similar procedure as that for query state tokens. In each environment, we pass the entire in-context dataset $\mathcal{D}$ to the model. We then iterate through $t=[T, T-2, \dots, 1]$ and collect representations from the model in response to token $\mathcal{D}_t = (s_t, a_t, s'_t, r)$ if the transition $(s_t, a_t, s'_t, r)$ has not already been collected for this environment. We work backwards under the assumption that model representations are more rich as in-context experience increases, and thus more likely to contain task-relevant variables. For each token that produces a regressor, we define variables of interest relative to $s_t$ (e.g., value function for state $s_t$). We did not see a difference when we defined the variables we tested relative to $s_t'$ instead.

\section{Gradient Attribution Method}
\label{sec:appendix-linear-decoder}
To get gradient attributions, we use integrated gradients \citep{sundararajan2017axiomatic}. As a reminder, the model output is a vector defining weights over actions. We calculate the gradient of the model's output for the optimal action with respect to input tokens. We define the baseline inputs as the original context memory dataset $\mathcal{D}$ but with actions ablated (that is, $a=\textbf{0}$). We integrate over $20$ steps.

\newpage
\section{Tests for Model-Free Reasoning}
\label{sec:appendix-model-free}
\begin{figure}[h!]
  \centering
  \includegraphics[width=\textwidth]{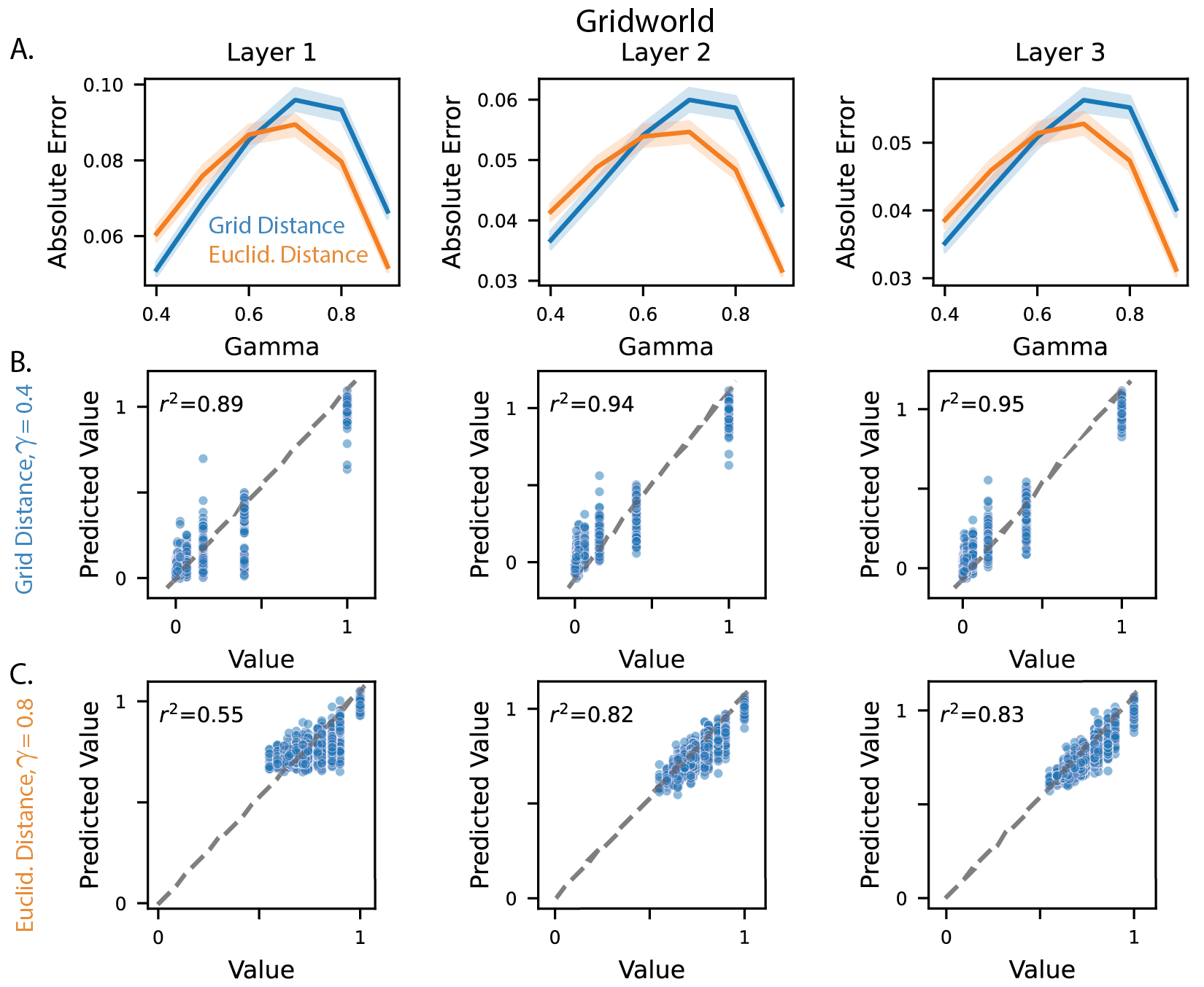}
  \caption{Tests for model-free reasoning in gridworld. \textbf{A.} Absolute error in test set for decoders fit on $V^*$ (Grid Distance, blue) and $V_e$ (Euclid. Distance, orange) across different values of $\gamma$. \textbf{B.} Predicted value vs actual value in test set, for $V^*$ and $\gamma=0.4$. \textbf{B.} Predicted value vs actual value in test set, for $V_e$ and $\gamma=0.8$.}
  \label{fig:19}
\end{figure}

As a probe for model-free reasoning, we tested whether or not value information could be decoded from model representations. Specifically, we test whether, at decision time, the model utilizes value information of the query state to drive decisions.

Specifically, we assess whether the model encodes value estimates $V^*(s) = \mathbb{E}_{\pi} \left[ \sum_{t=0}^{\infty} \gamma^t R_{t+1} \mid S_0 = s \right]$ for its current state $s$ under an optimal policy.
We trained linear decoders on query token representations to predict $V(s)$, but overall did not find evidence that a value gradient could be extracted from model representations (see Apps. \ref{sec:appendix-linear-decoder} and \ref{sec:appendix-model-free}).
In gridworld, $V(s)$ can be decoded with high accuracy (Appendix~\ref{sec:appendix-model-free}).
However, decoding is more accurate when $V(s)$ is defined in terms of Euclidean distance to the goal, rather than over the true 4-dimensional action space (Appendix~\ref{sec:appendix-model-free}).
This suggests that the model encodes spatial structure rather than true value gradients—its apparent $V(s)$ reflects geometric regularities, not action-contingent reward prediction.
In tree mazes, decoded value estimates are localized: $V(s)$ is only reliable within 2--3 steps of the reward (Appendix~\ref{sec:appendix-model-free}).
This narrow value gradient is insufficient to guide behavior over the full task horizon.

We start with fitting linear decoders in gridworld. Let $s$ be a query state and $s_{goal}$ be the reward state. The variable we predict from the model representations is $V^*(s) = \mathbb{E}_{\pi} \left[ \sum_{t=0}^{\infty} \gamma^t R_{t+1} \mid S_0 = s \right]$ for state $s$, taking an optimal policy. Equivalently, $V^*(s) = \gamma^{d(s, s_{goal})}$, where $d(s, s')$ describes the number of actions needed to navigate from $s$ to $s'$. Thus, $V^*$ describes an exponentially decaying value gradient in terms of action distance. To evaluate the decoder, we plot the test error against the value function $\gamma$ (Fig \ref{fig:19}A). We note that, although the lowest error is achieved at $\gamma=0.4$ (Fig \ref{fig:19}B), this is not actually a useful parameterization for a value function as the value gradient decays quickly for states more than 2 steps away from reward. However, the test error at $\gamma=0.8$ is as low as $0.04$ in the final model layer.

Given that model representations capture the environment structure well, we suspect that the high decoding accuracy for $V^*$ may result from the spatial organization of representations. That is, if XY location information is contained in representations, a linear decoder could also do fairly well at predicting $V^*$. As a control, we define $V_e = \gamma^{d_{e}(s, s_{goal})}$ where $d_{e}(s, s')$ describes the Euclidean distance from $s$ to $s'$. We find decoding error is lower for $V_e$ than for $V^*$. However, $V_e$ does not reflect the actual action affordances in gridworld (since action space is only up/right/left/down). Thus, we conclude that the strategy used by the model may have more to do with learning the latent Euclidean structure of the environment than learning a value function across action space (as would be expected in standard model-free algorithms).

\begin{figure}[h!]
  \centering
  \includegraphics[width=\textwidth]{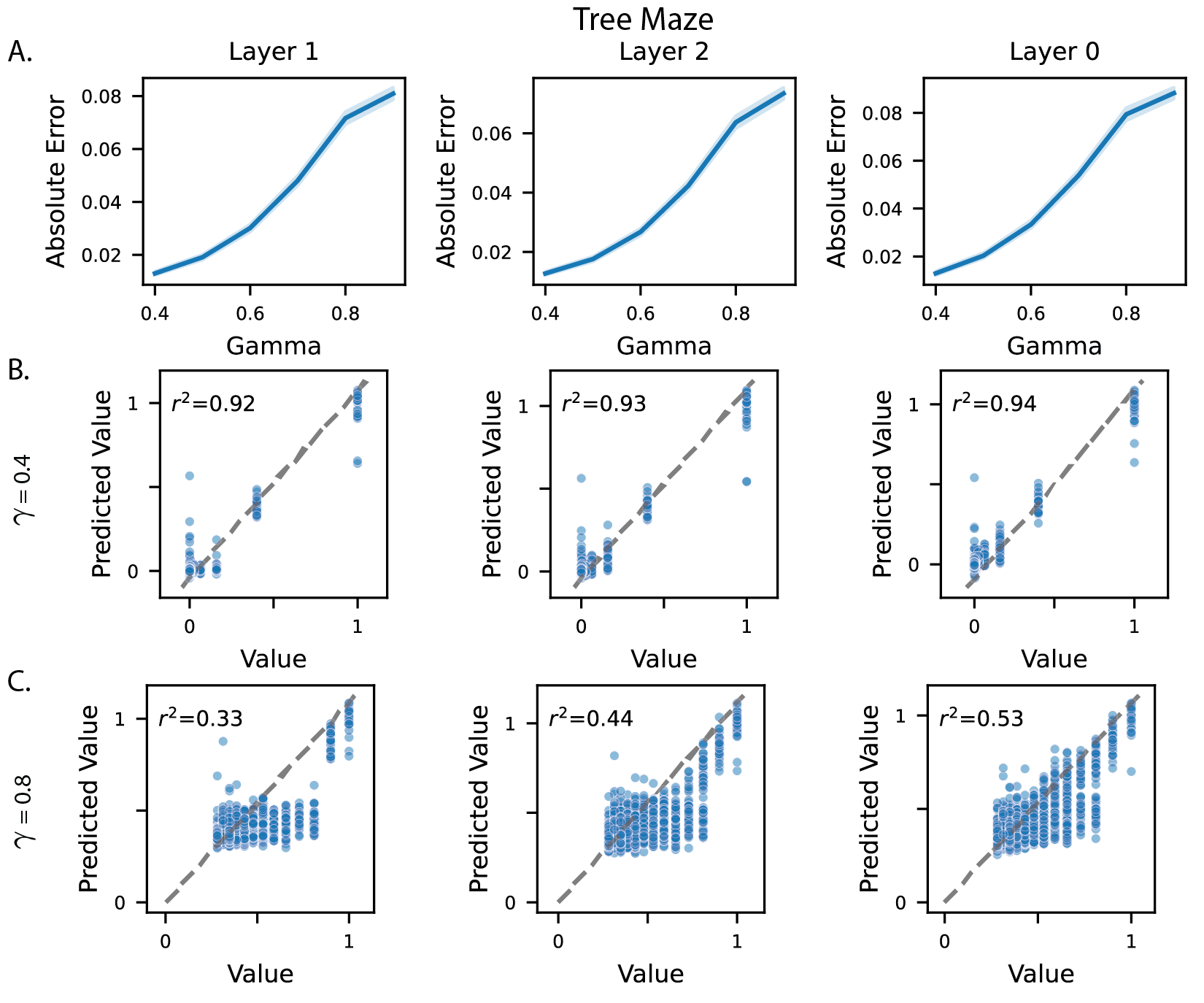}
  \caption{Tests for model-free reasoning in tree mazes. \textbf{A.} Absolute error in test set for decoders fit on $V^*$ across different values of $\gamma$. \textbf{B.} Predicted value vs actual value in test set, for $V^*$ and $\gamma=0.4$. \textbf{B.} Predicted value vs actual value in test set, for $V^*$ and $\gamma=0.8$.}
  \label{fig:20}
\end{figure}

We repeat this analysis in the tree maze task. We define $V^*$ as before and plot the test error against the value function $\gamma$ (Fig \ref{fig:20}A). We find that decoding error increases with $\gamma$. We plot predicted $V^*$ vs actual $V^*$ for the lowest and highest $\gamma$ values (Fig \ref{fig:20}BC). We find that at $\gamma=0.4$, $V^*$ is well fit, however the value gradient is only meaningful for states that are 1-2 steps away from reward. Thus, this is likely not useful as a model-free RL signal. Conversely, at $\gamma=0.8$, the decoding does not perform well, and at most reaches $r^2=0.53$.

\newpage
\section{Tests for Model-Based Reasoning}
\label{sec:appendix-model-based}
\begin{figure}[h!]
  \centering
  \includegraphics[width=\textwidth]{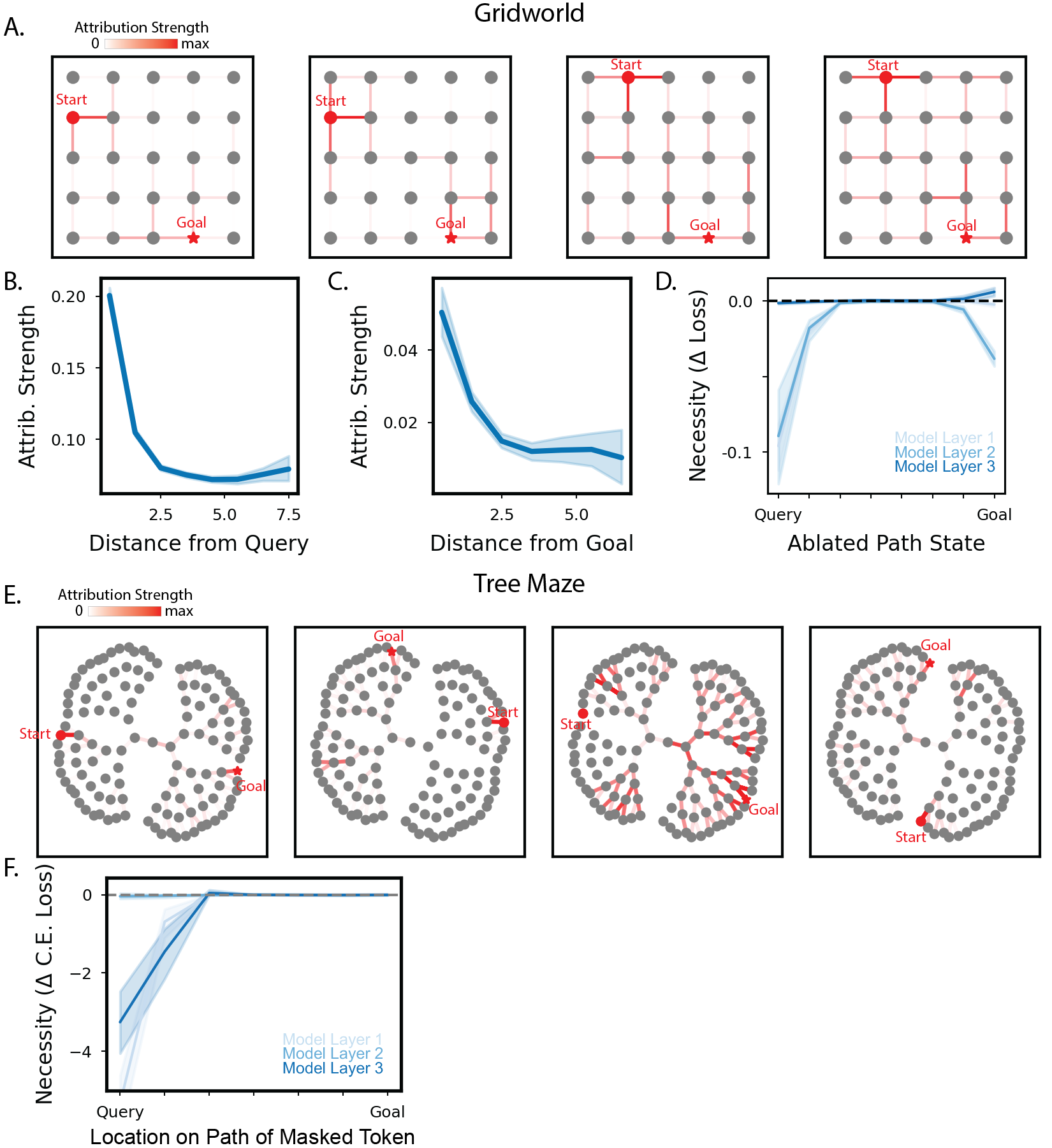}
  \caption{Tests of model-based reasoning. \textbf{A.} As in Fig \ref{fig:5}A, but for four example gridworld environments. \textbf{B., C.} As in Fig \ref{fig:5}BC, but for gridworld tasks. \textbf{D.} Measurement of necessity for each context memory token on the path from the query state to the goal state. We measure the change in cross--entropy loss when the token is ablated, and we plot this against the location of the token. Tokens are ablated by masking the query-to-token attention at the desired model layer. In all tests, the query state remains the same. Line color indicates which layer of the model the intervention was conducted in. We show average value across $50$ environments, with $95\%$ confidence interval shading. Since there are multiple possible paths from query token to goal in gridworld, we define the path as the sequence of states the agent would have taken had we allowed it to navigate to reward. We include only cases where the model successfully navigates to goal. \textbf{E.} As in Fig \ref{fig:5}BC, but for four more tree maze examples. \textbf{F.} As in (D) but for the tree maze task.}
  \label{fig:21}
\end{figure}

We next probe for signatures of model-based reasoning. That is, we look for evidence that the model utilizes path planning to choose the correct action from the query state. This is connected to questions of state tracking \citep{li2025language} and understanding how models simulate successive transitions between states. \citet{li2025language} propose different ways that transformer models can do this path planning, from forward rollouts to more sophisticated, mergesort-like algorithms. Each of these algorithms require simulating transitions through intermediate states between query and goal. Thus, to test for the presence of path planning, we look for evidence that information about intermediate states are utilized at decision time. Specifically, we isolate decision time as computations conducted in the query token stream.

We first begin with gridworld and analyze the gradient attributions over the input context tokens (which are themselves transitions). We plot individual examples of these attribution maps and summary statistics in Fig \ref{fig:21}A-C. Taken together, it does not appear that the model relies on path planning from the current state to the goal. We further test with ablations of states on the path from query to goal. Specifically, at each model layer we conduct a necessity test where we mask attention from the query token to context tokens containing the ablated state. We then measure how attention ablations impact the original cross-entropy loss (Fig \ref{fig:21}D). We find that ablating intermediate states does not impact cross-entropy loss.

We conduct the same analyses in tree maze (Fig \ref{fig:21}EF) and find similar results. We conclude that path planning as done in typical model-based reasoning is not a strategy that the model is relying on to solve either tasks.

Our conclusion contrasts with prior work that analyzed the behavior of meta-learned RL agents and suggested that they implement a form of model-based planning \citep{wang2016learning, ritter2018been}. It is possible that meta-learning can discover more typical model-based strategies in other settings, and these discrepancies demonstrate that the mechanistic perspective taken here can provide insights into understanding meta-learned algorithms.

\newpage
\section{Gridworld Mechanisms}
\label{sec:appendix-gridworld-mech}

\begin{figure}[h!]
  \centering
  \includegraphics[width=\textwidth]{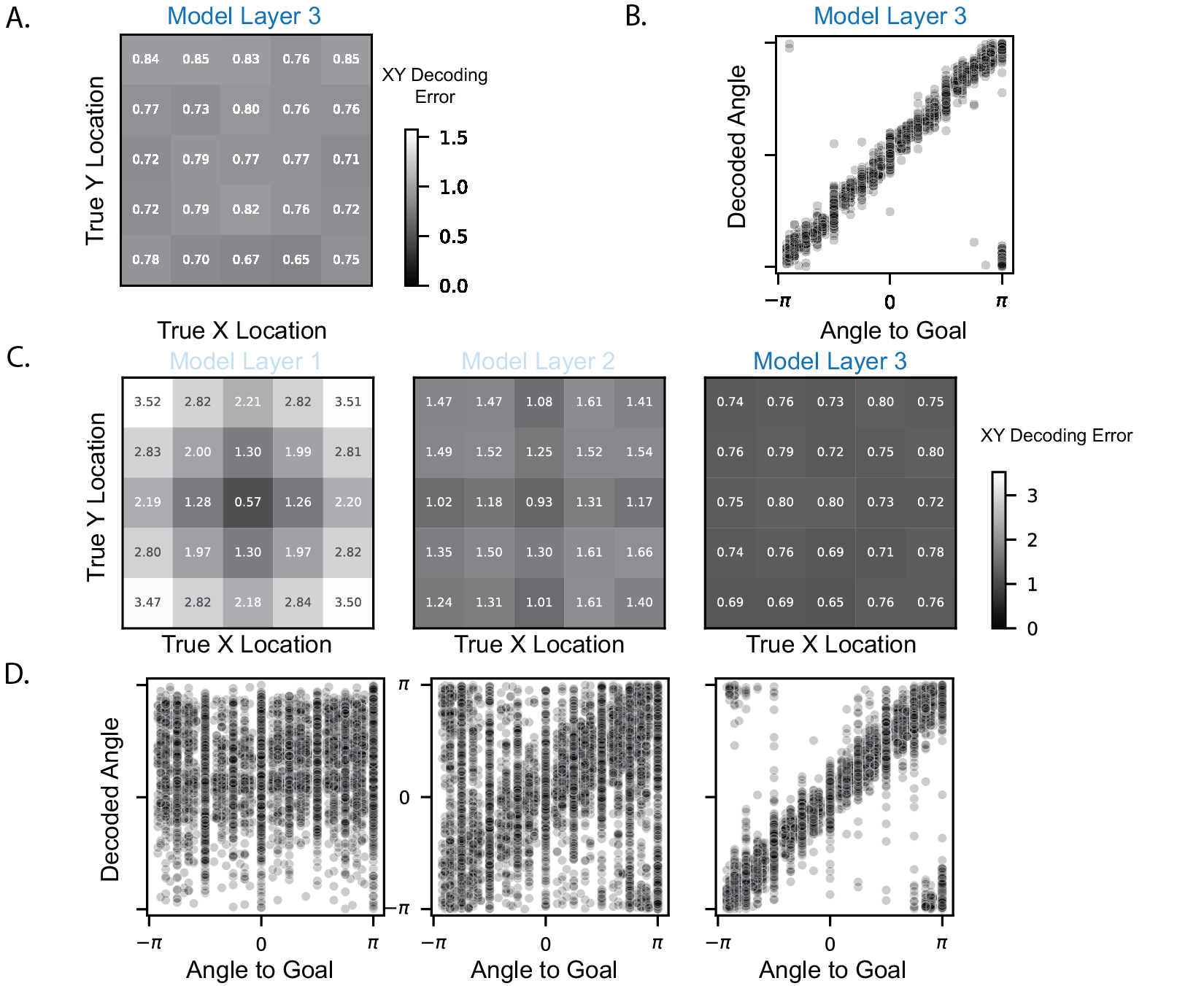}
  \caption{Additional decoding plots for gridworld task. \textbf{A., B.} As in Fig \ref{fig:6}BC, but for model layer 3. \textbf{C.} As in Fig \ref{fig:6}B, but decoding analysis is run on the context memory tokens that enter each layer of the model. \textbf{D.} As in Fig \ref{fig:6}C, but decoding analysis is run on the context memory tokens that enter each layer of the model.}
  \label{fig:22}
\end{figure}

In Figure \ref{fig:22}, we show additional decoding results for gridworld. In the main text, we discussed how XY location and angle to goal can be decoded clearly from the query token stream of the model. We also find that these variables can be decoded from the context tokens. We find this interesting as it connects to our findings in tree mazes where memory tokens contain not just information about the original event (i.e., transition), but also additionally computed features.

In results section \ref{sec:mechanisms}-Gridworld, we gave brief descriptions of a few analyses we ran. Here we will give more details on these analysis. First, we discussed how we showed that model performance relies on attending to tokens near the query and goal states in layer 2 (Fig.~\ref{fig:6}D).
We do this by masking attention from the query token to individual context tokens, following the ablation procedure described in Section~\ref{sec:not-mf-mb}.
We find that layers 1 and 3 are robust to these ablations, but performance degrades in layer 2 when attention to tokens near the query or goal is removed (Fig.~\ref{fig:6}D).

We also discussed how the attention patterns between context memory tokens shift from localized to distributed across model layers (Fig.~\ref{fig:6}), suggesting that the model first stitches transitions locally before constructing global structure.
To arrive at this conclusion, we analyze the spatial locality of attention between context-memory tokens to test whether transitions are integrated locally or globally.
Specifically, we plot attention strength as a function of spatial distance between token pairs (Fig.~\ref{fig:6}E).
We restrict this analysis to the first two layers, since context-to-context attention in the final layer does not influence the policy output.

\newpage
\section{Tree Maze Mechanisms}
\label{sec:appendix-tree-maze-mech}

\begin{figure}[h!]
  \centering
  \includegraphics[width=\textwidth]{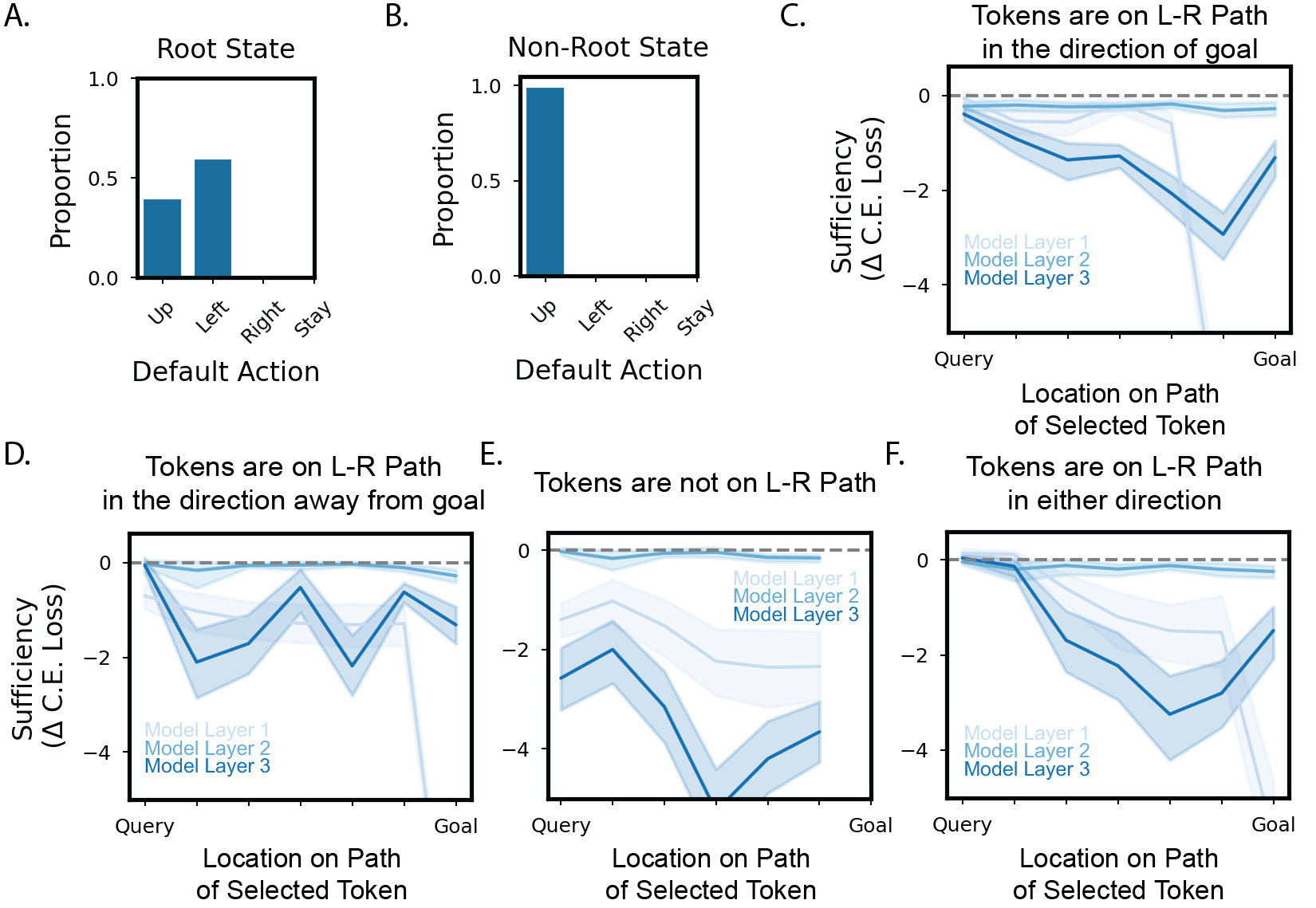}
  \caption{Additional analyses for tree maze task. \textbf{A.} Across 50 environments where reward information is ablated from the context, the proportion of each action taken by the model when at the root state. \textbf{B.} As in (A), but for non-root states. \textbf{C.} As in Fig \ref{fig:7}C, but we further restrict the context memory tokens to be on the L-R path and transitioning in the direction towards goal. \textbf{D.} As in (C), but transitioning in the direction away from goal. \textbf{E.} As in (C), but for context memory tokens that are non the L-R Path. That is, for the state indicated on the x-axis, we select context memory tokens that involve that state but are not on the L-R path. There are no transitions involving the goal that are not on the L-R path, and thus no data at that point. \textbf{F.} As in (C), but we restrict context memory tokens to be on the L-R path and do not further restrict by their directionality to or from the goal.}
  \label{fig:23}
\end{figure}

In Figure \ref{fig:23}, we show additional results for our tree maze analysis. In the tree maze task, the optimal action is often to transition to the parent node (specifically, this is true in all but the $6$ states that comprise the L-R path). This bias is reflected in the model. Without reward information, the model defaults to transition towards the parent node unless it is at the root node (Fig \ref{fig:23}AB). Thus, we believe the model takes its default action unless it accumulates enough evidence through its layer computations to do otherwise. As discussed in the main text, we think this is done by tagging context tokens on the L-R path.

 We also give more information here about analyses that we briefly described in results section \ref{sec:mechanisms}-Tree Maze. We discussed a hypothesis where, at decision time, the model tests if there are context-memory tokens that contain the query state and are tagged as being on the L-R path. If so, then the correct left/right action can be inferred from the same tagged tokens (in particular since inverse actions are also encoded).
 
 We find further evidence for this strategy by re-doing our sufficiency analysis from Fig \ref{fig:7}C with three additional restrictions: (1) tokens must be on the L-R path in the direction to the goal, (2) tokens must be on the L-R path in the direction away from the goal, or (3) tokens are not on the L-R path at all. We find that in the first two cases the model output is unaffected, but in the third case the output is negatively impacted ((Fig \ref{fig:23}C-F)). As long as the query token in the last layer can attend to context tokens that (1) themselves contain the query token and (2) are on the L-R path going towards or away from goal, the model output is unaffected (Fig \ref{fig:23}C-F). The results of these perturbations are consistent with our hypothesis that decision-making in the model relies on identifying if the query state is on the L-R path via intermediate computations stored in context-memory tokens.

\end{document}